\documentclass[a4paper,fleqn]{cas-sc}

\usepackage[numbers]{natbib}

\usepackage{amssymb}
\usepackage{amsxtra}
\usepackage{amsmath}
\usepackage{amsthm}
\usepackage{mathrsfs}
\usepackage{tabularx}
\usepackage{threeparttable}
\usepackage{subfigure}
\usepackage{multirow}
\usepackage{multicol}
\usepackage{xcolor}
\usepackage{transparent}
\usepackage{makecell}
\usepackage{algorithm}
\usepackage{booktabs}
\usepackage{hyperref}
\usepackage[capitalise]{cleveref}
\usepackage{enumitem}
\usepackage{algpseudocode}
\usepackage{xcolor}
\usepackage[normalem]{ulem}
\usepackage{subcaption}

\begin{document}

\let\WriteBookmarks\relax
\def\floatpagepagefraction{1}
\def\textpagefraction{.001}

\title [mode = title]{
    Intuitive Axial Augmentation Using Polar-Sine-Based Piecewise Distortion for Medical Slice-Wise Segmentation
}
\shorttitle{Axial Distortion}
\shortauthors{Yiqin, et~al.}

\author[1]{Yiqin Zhang}[orcid=0000-0003-2099-2687]
\ead{zyqmgam@163.com}
\ead[url]{https://github.com/MGAMZ}
\credit{Project administration, Conceptualization, Data curation, Formal analysis, Investigation, Methodology, Software, Resources, Validation, Visualization, Writing - Original draft preparation, Writing - review and editing}

\author[1]{Qingkui Chen}
\cormark[1]
\ead{chenqingkui@usst.edu.cn}
\credit{Project administration, Funding acquisition, Supervision, Resources, Writing - Reviewing and Editing}
\cortext[1]{Corresponding Author's Email: \url{chenqingkui@usst.edu.cn}}

\author[3]{Chen Huang}
\credit{Resources, Supervision}

\author[2]{Zhengjie Zhang}
\credit{Data curation, Formal Analysis, Investigation}

\author[1]{Meiling Chen}
\credit{Formal Analysis, Validation, Visualization, Writing - Reviewing and Editing}

\author[1]{Zhibing Fu}
\credit{Writing - Reviewing and Editing}

\affiliation[1]{
    organization={University of Shanghai for Science and technology},
    addressline={JunGong Road 516, YangPu District}, 
    city={Shanghai},
    country={China}
}
\affiliation[2]{
    organization={Huashan Hospital, Fudan University},
    city={Shanghai},
    country={China}
}
\affiliation[3]{
    organization={Shanghai General Hospital, Jiaotong University},
    city={Shanghai},
    country={China}
}

\begin{abstract}
Most data-driven models for medical image analysis rely on universal augmentations to improve accuracy. Experimental evidence has confirmed their effectiveness, but the unclear mechanism underlying them poses a barrier to the widespread acceptance and trust in such methods within the medical community. We revisit and acknowledge the unique characteristics of medical images apart from traditional digital images, and consequently, proposed a medical-specific augmentation algorithm that is more elastic and aligns well with radiology scan procedure. The method performs piecewise affine with sinusoidal distorted ray according to radius on polar coordinates, thus simulating uncertain postures of human lying flat on the scanning table. Our method could generate human visceral distribution without affecting the fundamental relative position on axial plane. Two non-adaptive algorithms, namely Meta-based Scan Table Removal and Similarity-Guided Parameter Search, are introduced to bolster robustness of our augmentation method. In contrast to other methodologies, our method is highlighted for its intuitive design and ease of understanding for medical professionals, thereby enhancing its applicability in clinical scenarios. Experiments show our method improves accuracy with two modality across multiple famous segmentation frameworks without requiring more data samples. Our preview code is available in: https://github.com/MGAMZ/PSBPD.
\end{abstract}

\begin{graphicalabstract}
\includegraphics[width=\textwidth]{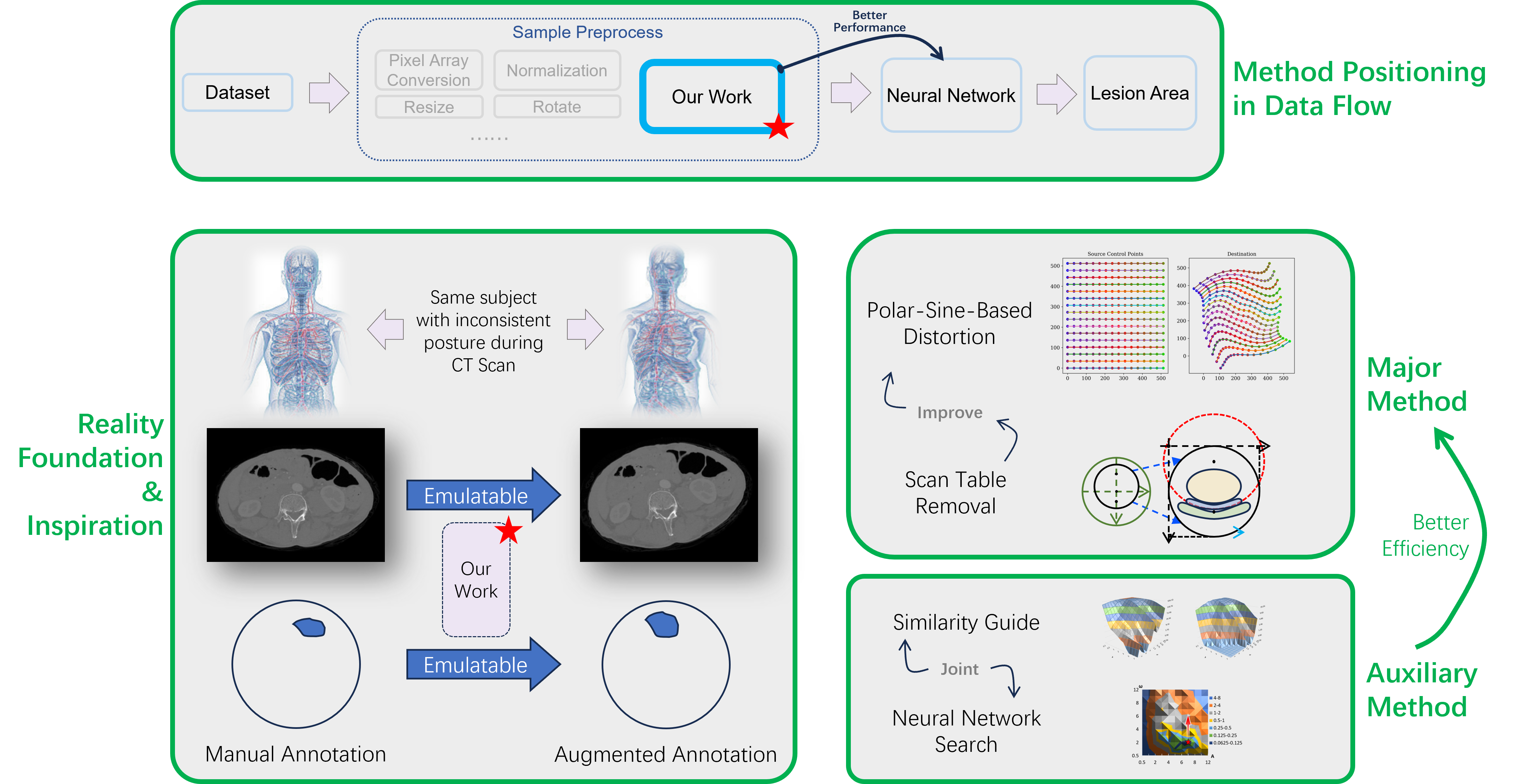}
\end{graphicalabstract}

\begin{highlights}
    \item An plug-and-play augmentation method for medical radiologic imaging on axial plane.
    \item Inspired by and can simulating patients' posture uncertainty during the scan.
    \item Introduced DICOM Meta-Data-Driven geometric modeling
    \item Similarity-Guided training strategy.
\end{highlights}


\maketitle
\setlength{\parskip}{\baselineskip}

\section{Introduction}
\label{sec:intro}

\textbf{Research Significance.} Deep learning models have seen significant advancements in healthcare, including the widespread adoption of automated lesion segmentation, which alleviated the workload of radiologists and contributed to diagnostic precision. The up-to-date data-driven models often require a sufficient amount of data to obtain acceptable results, which can lead to unaffordable research costs in the medical field. In the computer vision community, various data augmentation techniques are widely used as part of preprocessing \citep{BERNAL201964} to alleviate data scarcity. 
In the field of medical imaging, an excellent augmentation method should meet several requirements: \textbf{1)} Reduce the annotation cost of medical datasets. \textbf{2)} Fully tap the value of existing precious annotations without the need for more radiologists. \textbf{3)} Being easily comprehensible to doctors in clinical scenarios, thus ensuring practical applicability \citep{Reddy10134984, Hatherley_Sparrow_Howard_2022}. Considering there’re plenty of existing neural networks for medical image analysis, we argue that creating a methodology that is applicable to most models is more impactful than solely focusing on developing an enhanced neural network model.

\textbf{Traditional Augment: Effective but Counterintuitive.} Recent research \citep{GARCEA2023106391} suggests that some data augmentation methods transform the original data in more extreme ways (e.g.  Erasing, Solarize, cross-domain synthesis, and generative models, etc..). Medical experts often find these samples unreasonable, not present in the real world, and devoid of practical significance. Samples generated through these techniques may not conform strictly to anatomical standards. For example, in a CT scan sequence, if the HU value for the heart region is set to a constant and the locations of the bladder and liver are interchanged, such samples are generally classified as noise and are removed during the standard data analysis procedures \cite{Liu10230495,SUN2025102741}. However, the reality is not the case as to deep learning. Experimental results show that neural networks seem to still prefer these exaggerated enhancement methods, learning more knowledge from the “unreasonable” samples and showing better accuracy on downstream tasks. The reasons for this phenomenon are controversial, but it is clear that the interpretability of these data augmentation methods is very poor, which is particularly important in medical AI scenarios \citep{salahuddin2021, VANDERVELDEN2022102470}.

Physicians typically scrutinize how data is collected and its quality. If they find out a model has been trained on deliberate “noise” data and claims a promising outcome, they’re likely to be skeptical. It’s counterintuitive to deliberately introduce noise for better accuracy, particularly when we can’t clearly justify “why it can.” As a result, enhancing the level of interpretability and optimizing the system’s transparency contribute to the potential application of a method in clinical scenarios.

\begin{figure}[tbp]
    \centering
    \includegraphics[width=\linewidth]{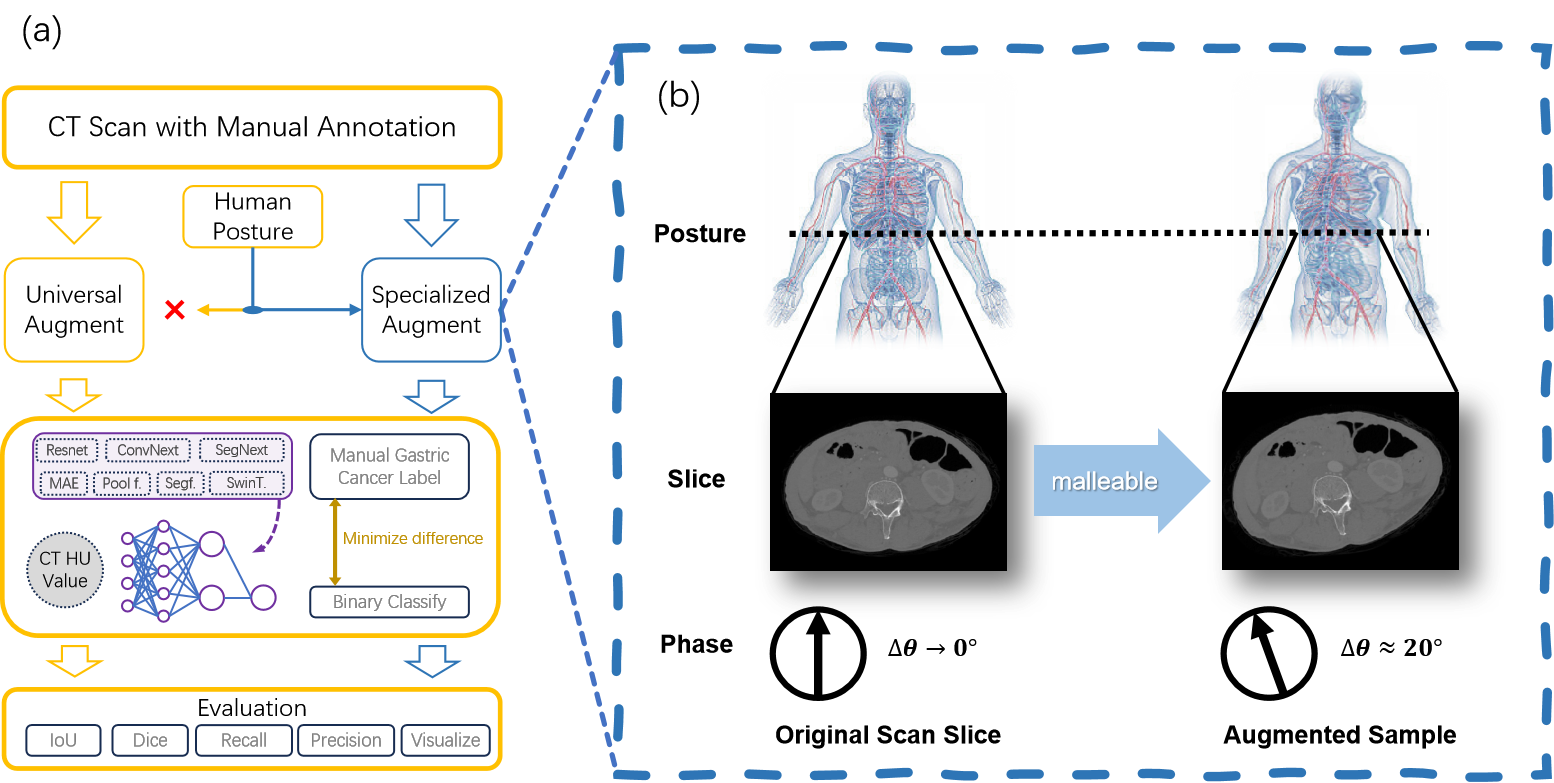}
    \caption{Overview of the proposed method.} (a) shows common frameworks tend to utilize universal augment method on CT dataset, without considering the effect of posture uncertainty. (b) is a pair of slices from the same subject with different posture, leading to variance on slices. Our augmentation method is able to simulate this malleable variance, thus generating more samples for neural networks.
    \label{fig:AugmentOverview}
\end{figure}

\textbf{Medical Radiologic Scan Sequences: Unexplored Potential.} We notice there’re significant differences in the physical imaging processes between radiologic scan images and common camera images. Patients undergoing these radiologic scans are usually required to lie flat on their backs with their hands behind their heads and arms raised above their heads for the examination, and the posture adopted by the patient is difficult to control precisely. Their motions directly influence the reconstruction results on every slice \citep{Toth2012} \cref{fig:AugmentOverview}. This means that the same subject will obtain different reconstruction results when undergoing different scans, and these diverse samples can help neural networks fit more efficiently \citep{MUMUNI2022100258}. However, due to limitations in medical resources and the harmful effects of radiation, it is not acceptable to repeatedly scan the same subject \citep{Messmer2007, wade2007ethics}. Considering that the subject itself does not change in the aforementioned differences, it is possible to simulate the reconstruction slice with human posture changes from the known scan sequences. Moreover, standard DICOM files potentially enables the precise execution of various data preprocessing tasks through the utilization of non-adaptive algorithms.

As previously noted, simulation algorithms developed in response to this observation offer enhanced interpretability. In clinical practice, it is expected that the same patient will yield varying reconstruction results across different scans. Consequently, the samples produced by these methods are deemed to align with clinical realities and do not introduce any inexplicable elements. It is a well-acknowledged fact within the radiology community that radiologists often encounter studies that include multiple series, with each series displaying similar yet subtly different human body postures \citep{MARSHALL2022271,Bali2023}. Medical professionals will readily comprehend the enhancement in accuracy achieved through the use of these samples. Because this is analogous to gathering additional data on a patient, which inherently aids in the intuitive understanding and facilitation of any data analysis process.

\textbf{Our Method: Intuitive Distortion.} According to the above, we proposed a Polar-sine-based Piecewise Affine Distortion specifically for medical radiologic image augmentation. Our approach, maps the reconstruction results of the original cross-sections to a distribution of which containing the subject’s posture changes. This is done to simulate the appearance when the body of the scanned individual exhibits slight distortions. In order to prevent disrupting the fundamental relative positional relationships of the human body's tissues during augmentation, we construct a random transformation algorithm based on polar coordinates combined with sine functions. The scan table will be precisely identified and eliminated utilizing metadata-driven geometric algorithm. The elimination helps to the undesirable artifacts introduced by the augmentation. To align AI more closely with the intuitive aspects of medical practice and help reduce the cost when applying the augmentation to a new dataset, we incorporate similarity metrics to regulate the intensity of distortion, preventing unreasonable augmented samples. This Non-Neural approach also improves the efficiency of neural network search, thereby conserving resources during model fine-tuning.

In order to validate our methodology, we assembled a high-quality dataset of gastric cancer CT scans, which are collected over a period of seven years during clinical practice. This dataset is meticulously annotated by two medical professionals, resulting in the creation of a few-anno CT dataset comprising 689 studies. In order to ensure the reproducibility and compatibility of our proposed method, we incorporated two public datasets and train from scratch on seven distinct models across two paradigms. The empirical evidence indicates that our method not only maintains a satisfactory level of interpretability but also improve the precision across the majority of applicable scenarios.

Our preview code is available in: https://github.com/MGAMZ/PSBPD. In summary, our augmentation algorithm has the following features:

\begin{itemize}[label=\textgreater] 
    \item It can generate any number of augmented sequences with differences from a single scan sequence, with controlable impact on the crucial relative positional features of human organs to align with anatomical intuition, thus increasing the potential of applications within clinical scenarios.
    \item The application of similarity metrics facilitates a swift determination of the optimal parameter spectrum, thereby augmenting the deployment efficacy on diverse datasets.
    \item It can eliminate the noise introduced by the scan table by using DICOM-metadata-based geometric positioning, achieving both excellent effects and speed.
\end{itemize}

\section{Related Works}

\textbf{Medical Radiologic Image Preprocess.} \citep{Chen7789988} and \citep{Lartaud9321056} jumped out of the traditional idea of improving the accuracy of the model, and matched the physical imaging process of CT into the neural network, which greatly improved the contrast of images and the visibility of some difficult to observe tissues. This method is highly interpretable, but requires a calibration for each CT imaging device to obtain certain required parameters. This reduces its ease of use and accessibility. \citep{Toth2012} emphasized the volumetric measurement of CT images and proposed a 2.5-D augmentation method. \citep{Liu10230495} skillfully isolated the image of the lesion area and then altered its background, which could even come from irrelevant samples. This approach aligns with deep learning practices and experiences \citep{Kumari10455714, chlap2021review}, yet it offers little in terms of explainability, which is crucial for medical applications. For instance, it is highly unconventional for images of the stomach and kidney to be present within the same slice. The approach we are suggesting has no need for hardware calibration. Although certain procedures do require DICOM metadata, these are not mandatory.

\textbf{Affine in Medical Image Processing.} \citep{El10005544} reviewed the augmentation method to ease data scarcity of medical image. They pointed out that affine transformations (e.g., flip, rotation, translation, scaling, cropping and shearing) have widely used as a part of the pre-processing workflow for medical images. \citep{GARCEA2023106391} further analyzed the latest research on augmentation and believed that augmentation is very effective for medical datasets. \citep{ZHAO2023102786} used multiple Affine Matrices on high-dimensional CT feature maps to differentially deform the vertebral bodies and surrounding soft tissues, leading to better registration accuracy. This is mainly due to the different elastic deformation characteristics between different tissues inside the body. The method aligns with our thinking in assuming the human body’s non-rigid nature. However, our mapping is more granular and extends beyond just local tissue structures. \citep{ZHANG2023120303} proposed an affine-enhanced arterial spin labeling (ASL) image registration method for MRI images. In this method, the affine transformation will be applied to image according to six parameters learned by deep learning neural network. \citep{Gu8803794} proposed a two-stage unsupervised learning framework for deformable medical image registration. This method features a larger granularity and integrates a complicated two-phase modeling strategy.

\section{Methods}

\subsection{Data Preprocess Overview}

\begin{figure}[tbp]
    \centering
    \includegraphics[width=\textwidth]{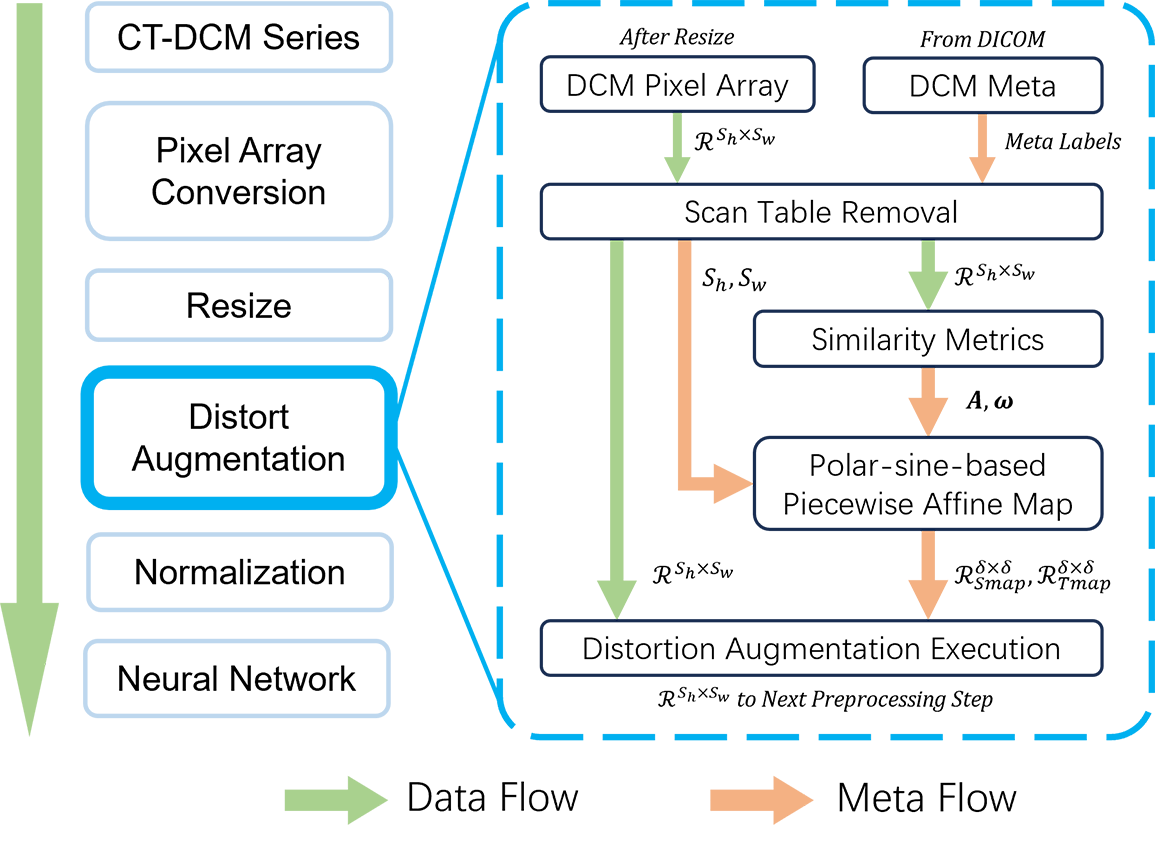}
    \caption{Overview of data preprocess.} Symbol definitions are available in Table 1. The bold blue module is the proposed method. This approach is distinct and separate from neural networks and other processing, offering significant compatibility with various training frameworks. In the context of data flow, the proposed method is positioned subsequent to size normalization and prior to normalization. The inclusion of Scan Table Removal is optional, serving to alleviate potential negative effects resulting from the proposed augmentation. Utilization of Similarity Metrics aids in estimating viable $A, \omega$, thereby enhancing the efficiency of parameter search.
    \label{fig:DataPreprocessOverview}
\end{figure}

According to the latest research \citep{Shi10310260, Zhang10105457, Pup10365170}, deep learning models need to preprocess the data with several augmentation before inputting it into the model. The method we proposed is one part of the preprocess, as indicated in \cref{fig:DataPreprocessOverview}. All symbols used in mathematical procedure description of the proposed distortion are shown in \cref{table:SymbolDefinitions}.
\begin{table}[tbp]
    \fontsize{8pt}{10pt}\selectfont
    \raggedright
    \caption{Usage and Style for Symbols.}
    \label{table:SymbolDefinitions}
    \begin{tabular}{ccc}
        \hline
        Symbol & Quantity & Range \\
        \hline
        $\theta$ & Polar Angle & $[0, 2\pi]$ \\
        $r$ & Radial Coordinate & $[0, +\infty)$ \\
        $\Theta$ & Pole/Origin Coordinate & $\mathbb{R}$ \\
        A & Amplitude & $[0, +\infty)$ \\
        $\omega$ & Angular Frequency & $[0, +\infty)$ \\
        $\phi$ & Initial Phase & $[0, +\infty)$ \\
        $\delta$ & Mesh Grid Dense & $n \in \mathbb{Z}^+$ \\
        $S_h$ & Array Height & $\mathbb{Z}^+$ \\
        $S_w$ & Array Width & $\mathbb{Z}^+$ \\
        $\Psi$ & Random Value & [0,1] \\
        $R^{S_h \times S_w}$ & Array with size $h \times w$ & $\mathbb{R}$ \\
        $R_{Smap}$ & Control Map Source Array & $\mathbb{R}$ \\
        $R_{Tmap}$ & Control Map Target Array & $\mathbb{R}$ \\
        $R_{Lin}$ & Linear Space Array & $\mathbb{Z}^+$ \\
        $\tau$ & Control Map Refresh Rate & $\mathbb{Z}^+$ \\
        $\mathcal{G}$ & Crt. to Pol. Cord. Convert & N/A \\
        $\mathcal{\varGamma}$ & Pol. to Crt. Cord. Convert & N/A \\
        $\mathcal{H}$ & Transformation Matrix & $\mathbb{R}$ \\
        \hline
    \end{tabular}
\end{table}

Pixel Array Conversion will read reconstructed image stored in dcm file series using method proposed in \citep{pydicom_v2.4.4}. Each patient’s scan includes multiple dcm files containing various meta data, which provides more possibilities for downstream tasks. The Resize operation is executed before the Distortion to achieve better accuracy, as the Distortion has an $O(H \times W)$ complexity. Normalization is performed after Distortion as a way to ensure that reconstructed pixel array conform to a standard distribution before being fed into the model.

\subsection{Implementation of Distortion}

\subsubsection{Affine Control Point Initialization}
To perform affine, we prepare a 2D grid of control points that are evenly distributed over the image. Control points function by establishing the mapping for a select few coordinates, thereby directing the transformation of the entire 2D matrix. The pixels corresponding to these control points remain constant before and after the mapping process, and they move in unison to the new location, adhering to the principles of classical piecewise affine mapping. During the affine process, the pixels corresponding to the control points are moved along with the control points themselves. So, the density of control point map determines the affine accuracy, more points lead to more independent affine operations. We use the linspace function to generate an evenly spaced square grid with $\delta^2$ control points:
\begin{equation}
    \begin{split}
        \mathcal{F}(a,b) =& Linspace(0,a,b) \\
        =& \{ a \cdot (i-1) / (b-1) \mid i=1,2,\cdot,b \} \\
        \mathcal{R}_{LinH}^\delta =& \mathcal{F}(S_w, \delta) \\
        \mathcal{R}_{LinW}^\delta =& \mathcal{F}(S_h, \delta) \\
        {\mathcal{R}_{Smap}}_{y, x}^{\delta \times \delta} = & ( {\mathcal{R}_{LinH}}_y^{\delta} , {\mathcal{R}_{LinS}}_x^{\delta})
    \end{split}
\end{equation}

\subsubsection{Affine Control Point Destination Calculation}

Based on the previous description, we should ensure the following two points in the mapping transformation: \textbf{1)} The continuity relationship between image pixels remains unchanged, and \textbf{2)} the distortion transformation should be reasonable comparing with actual scenarios, also conform to the distortion of the human body in reality.

To meet these requirements, we abandon the traditional calculation method based on the Cartesian coordinate system and convert the control point matrix to the polar coordinate system with the center of the $\mathcal{R}_{Smap}$ as the pole. For any radial line in the polar coordinate system, we distort it from a ray-like shape to a sine function shape, and map all points on this ray to its distorted version.
First, we randomly determine the actual distortion intensity parameters from a specified intensity range. This operation is to increase the intensity of data augmentation, since $A$ and $\omega$ controls the overall intensity of augmentation. We introduced a random factor $\Psi$, which can determine the amplitude $a$ and frequency $f$ actually applied in the transformation.
\begin{equation}
    a=\left(2\Psi-1\right)A, f=\left(2\Psi-1\right)\omega
    \label{eq:RandomAW}
\end{equation}

Then, correct the index order from pixel array space to physical location $X, Y$, and calculate the polar coordinates of the point with subscript indices $x, y$ in the polar coordinate with $\theta$ as the pole.
\begin{equation}
    \begin{split}
        y_{cord} =& \delta-y-1 \\
        r_{map},\vartheta_{map} =& \mathcal{G} \left(x,y,\Theta=\left(\delta / 2, \delta / 2\right)\right) \\
    \end{split}
\end{equation}

The key of the distortion is mapping each $\vartheta_{map}$ to a new location $\vartheta_{new}$. This conversion is performed with polar coordinate system, allowing us to easily control the absolute distance between each control point and the reconstruction center to remain constant, i.e. $(\frac{S_h}{\delta}, \frac{S_w}{\delta})$. This satisfies the second requirement shown at the beginning of this chapter. In actual scenario, human body's distortion amplifies where is close to the body's surface. In other words, a positive correlation between the distortion and the distortion center. When $\omega$ is not excessively large, the mapping of polar angles for all control points associated with a single polar angle tends to be comparable. Conversely, when W is extremely close to zero, the mapping resembles a rotational transformation, which barely impacts regional features.
\begin{equation}
    \vartheta_{new}=\vartheta_{map}+\frac{\pi}{8} \cdot a \cdot \sin{\left(\frac{r_{map}}{\delta} \cdot 2f\pi\right)}
\end{equation}

After conversion, we use $\mathrm{\Gamma}$ to invert the point $r_{img}, \vartheta_{new}$ to Cartesian coordinates, which means backspacing to pixel array space:
\begin{equation}
    \begin{split}
        x_{new},y_{new-cord} =& \mathcal{\varGamma} \left( r_{img},\vartheta_{new}, \Theta = \left(\frac{S_h}{2}, \frac{S_w}{2}\right) \right) \\
        y_{new} =& S_h - y_{new-cord} - 1 \\
    \end{split}
\end{equation}

Our method will apply this algorithm on all pixels to generate the target control map $\mathcal{R}_{Tmap}$ (6). The a and f parameters remain unchanged for one sample but varies across different samples. Obviously, a single transformation map only performs a fixed transformation, which does not conform to the idea of data augmentation, i.e. generating multiple different samples from one sample. The traditional rotate augmentation could also be achieved by adding a factor to $\vartheta_{new}$ (7).
\begin{equation}
    {\mathcal{R}_{Tmap}}_{y+1, x+1}^{\delta \times \delta \times 2}=(y_{new}, x_{new})
\end{equation}
\begin{equation}
    \vartheta_{new-rotate}=\vartheta_{new} + \mu, \mu \in \left( 0, \frac{\pi}{2} \right)
\end{equation}

\begin{figure}[tbp]
    \centering
    \includegraphics[width=\linewidth]{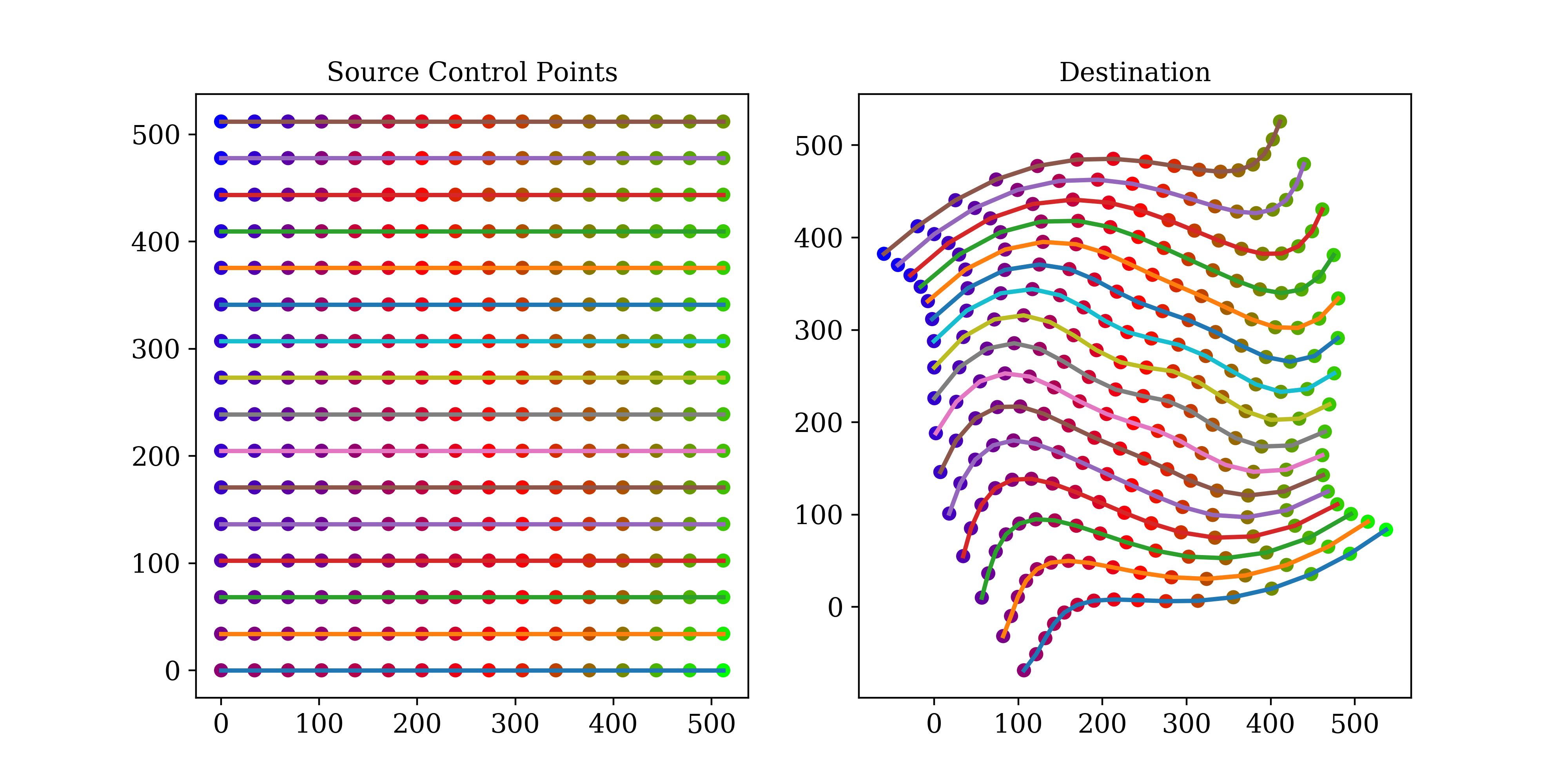}
    \caption{Generated control point map with $A=1$, $\omega=1$. The points of the same color correspond in two sub-figs. The positions of the converted control points have changed, with the points closer to the far end undergoing more drastic changes. However, the relative positional relationships between the control points remain unchanged, and points of similar colors still cluster together.}
    \label{fig:GeneratedControlPointMap}
\end{figure}

The algorithm's space and computational complexity are both $O(\delta ^2)$. A larger $\delta$ is advantageous for generating more precise images with segmented affine mapping. We consider $\delta \geqslant 16$ to make the graphics reasonable. These two points will be described in detail in the following sections. We give an example of generated control point map and its converted version in \cref{fig:GeneratedControlPointMap}.

\subsubsection{Piecewise Affine Execution}

Now that the affine control point map $\mathcal{R}_{Smap}$ and its destination map $\mathcal{R}_{Tmap}$ has been generated. Next, it is necessary to derive the pixel-level sampling relationship based on the mapping relationship of these control points, in order to sample a new matrix from the source matrix, which also means obtaining new samples. As to piecewise affine, a Delaunay triangulation of the points is used to form a mesh ${\mathcal{R}_\triangle}^{N_\triangle \times 3 \times 2}$ containing $i$ triangles. Delaunay triangulation function \citep{Dinas2014} is designed to maximize the minimum of all the angles of the triangles from a point set.
\begin{equation}
    {\mathcal{R}_\triangle}^{N_\triangle \times 3 \times 2} = Delaunay \left( {\mathcal{R}_{Smap}}^{\delta\times\delta} \right)
\end{equation}
where $N_\triangle$ is the number of the generated triangles of triangulation.

$Delaunay$ tends to avoid narrow triangles, as these triangles can lead to extreme distortions in image transformations. One triangle ${R_{\triangle, i}}^{3 \times 2}$ is composed of three control points $\left(\upsilon_1, \upsilon_2, \upsilon_3 \right) \in \mathbb{Z}^+$, $i$ is the index of $\triangle$. The piecewise affine will apply customized transformations for each triangle. We assume $\mathcal{H}_i^\triangle$ as one triangle's transformation matrix, and the piecewise affine can be described as \cref{eq:PiecewiseAffine}.
\begin{equation}
    \label{eq:PiecewiseAffine}
    {\mathcal{R}_{distorted}}^{S_h \times S_w} = \sum_{i = 1}^{n_\triangle} \mathcal{H}_i^\triangle R_{\triangle,i}
\end{equation}

\begin{figure}[tbp]
    \centering
    \includegraphics[width=\linewidth]{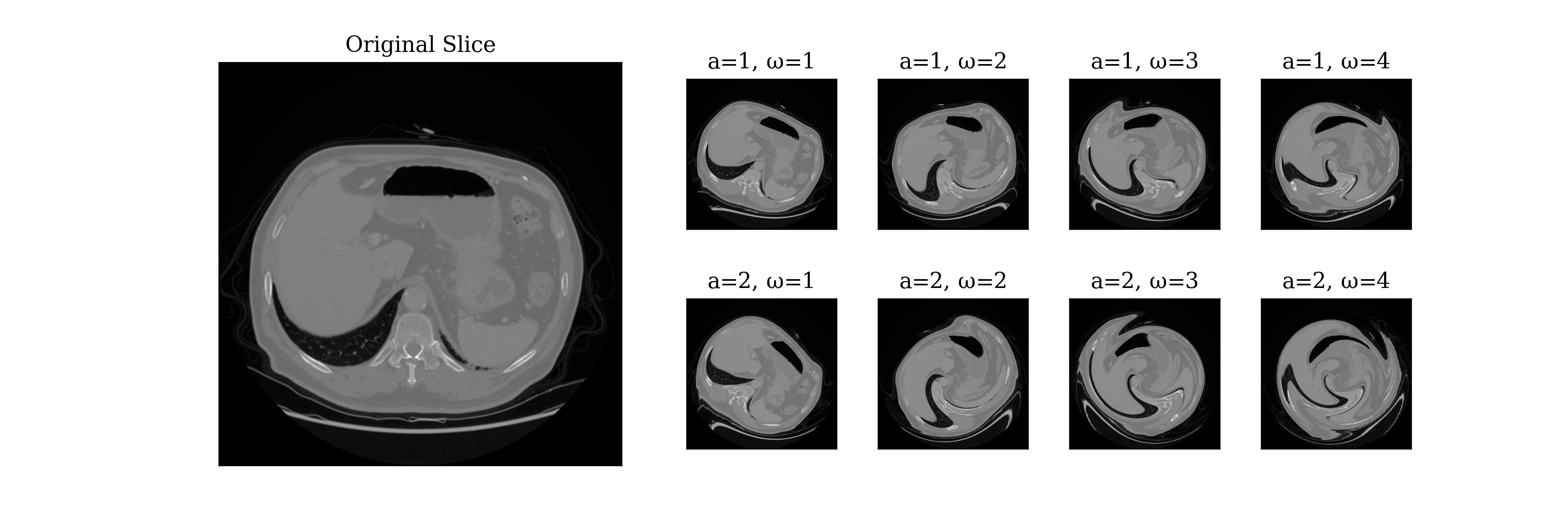}
    \caption{Distorted reconstructed slices with different parameters. $\Psi=1$ to better compare the effect. Larger $A$ and $\omega$ will lead to extreme distortion effect, but still remains continuity of adjacent areas. The scanning table is also distorted, this may introduce unnecessary noise.}
    \label{fig:DistortedDifferentParameters}
\end{figure}

We visualize examples using the proposed algorithm with $a \in \left\{1,2\right\}$, $\omega \in \left\{1,2,3,4\right\}$ in \cref{fig:DistortedDifferentParameters}.

\subsection{Metadata-driven Scan Table Removal}

In the usual course of CT reconstruction, the resulting sequences often include images of the CT Scan Table, predominantly positioned directly beneath the patient. Despite variations among subjects, the position of the CT Scan Table remains relatively constant. Consequently, the neural network, upon encountering this structure frequently in a large sample set and being informed by annotations that it is not a target of interest, tends to disregard the CT Scan Table and rarely misclassifies it as a relevant target. \cref{fig:DistortedDifferentParameters} illustrates not only the slice morphology under varying augmentation intensities but also highlights a concerning issue. The proposed augmentation method, which applies affine mapping to all pixels across the entire slice, transforms the CT Scan Table into a complex, multi-segment curve shape. Additionally, minor discrepancies in the Table’s position in the original space are exaggerated during the augmentation process, which can be interpreted as noise enhancement and is considered undesirable.

\begin{figure}[tbp]
    \centering
    \includegraphics[width=\linewidth]{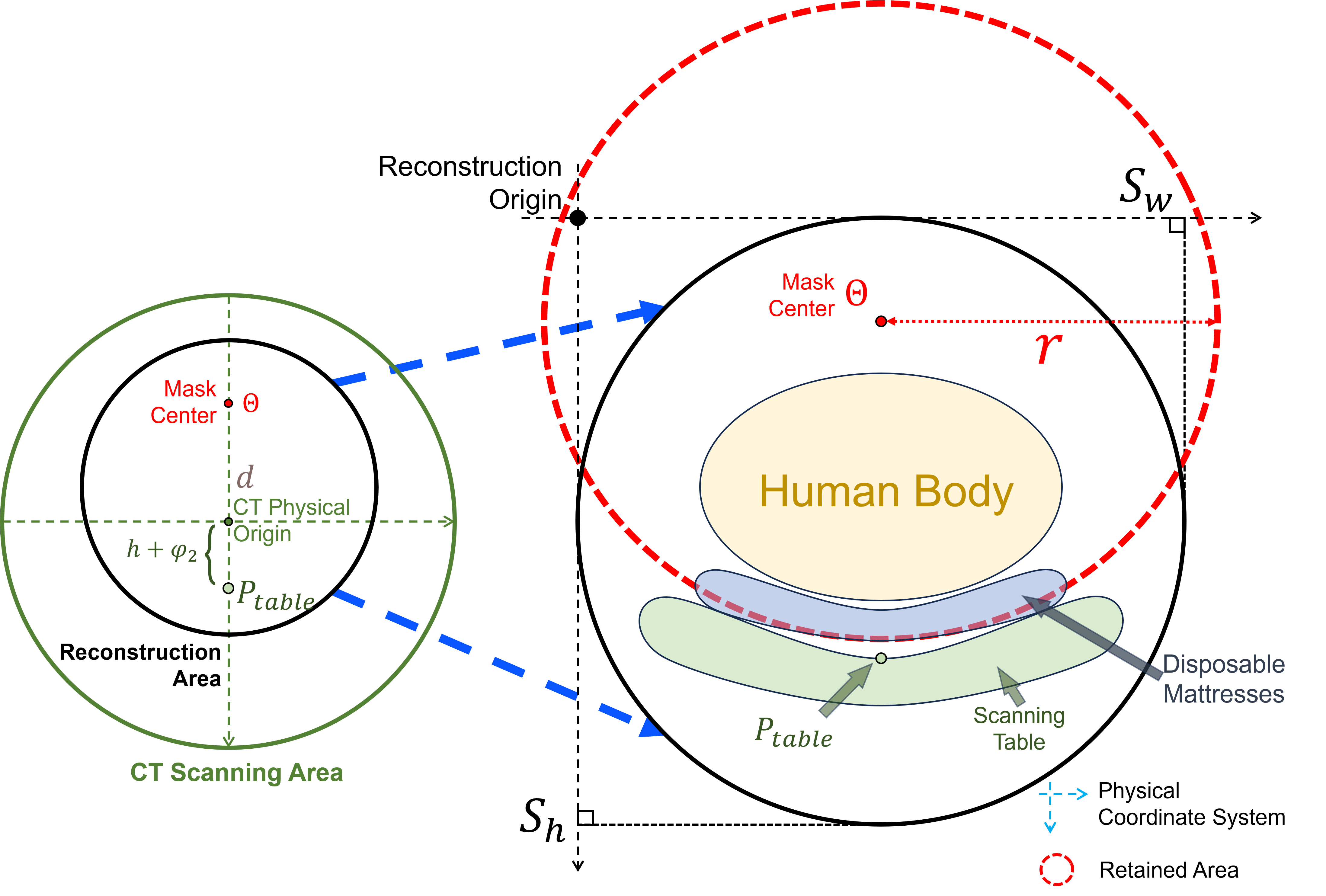}
    \caption{Geometric Positional Modeling used in scan table removal. The definition of DICOM symbols are available in \cref{tab:DICOM_Fields}.}
    \label{fig:GeometricModeling}
\end{figure}

The crux of this step lies in accurately identifying the pixels in each slice that correspond to the CT Scan Table. While a two-stage machine learning model that incorporates identification followed by segmentation is a viable approach, it entails a substantial increase in computational complexity. Capitalizing on the inherent benefits of medical sequence imaging, we employ DICOM metadata to spatially model the CT sequence, thereby precisely determining the location of the CT Scan Table. This method eschews the use of adaptive modules, thereby preventing an escalation in computational overhead. As a preparation, we deliberately preserved the DICOM metadata (while ensuring necessary anonymization) \citep{Gauriau2020}. All metadata and their symbols we use are illustrated in \cref{tab:DICOM_Fields}. The overview of geometric positional modeling is illustrated in \cref{fig:GeometricModeling}.
\begin{table*}[tbp]
    \fontsize{8pt}{10pt}\selectfont
    \centering
    \caption{DICOM fields used in our methods.}
    \label{tab:DICOM_Fields}
    \begin{threeparttable}
        \begin{tabularx}{\linewidth}{
            r
            >{\centering\arraybackslash}X
            >{\centering\arraybackslash}X
            >{\centering\arraybackslash}X
            >{\centering\arraybackslash}X
            >{\raggedright\arraybackslash}m{0.4\linewidth}
            }
            \hline
            Name & Req. Type \tnote{$\dagger$} & Symbol & DICOM Tag & Rep. Type & Describe \\
            \hline
            
            \makecell[r]{Reconstruction\\Diameter} & 3 & $S/2$ &\makecell[c]{G 0018\\E 1100} & DS & Diameter, in mm, of the region from within which the data is used in creating the reconstruction of the image. \\

            Table Height & 3 & $h$ & \makecell[c]{G 0018\\E1130} & DS & The distance in $mm$ of the top of the patient table to the center of rotation; below the center is positive. \\

            \makecell[r]{Image Position\\(Patient)} & 1 & \tnote{$\ddagger$} & \makecell[c]{G 0018\\E 1130} & DS & The $x, y$, and $z$ coordinates of the upper left hand corner (center of the first voxel transmitted) of the image, in $mm$. \\

            \makecell[r]{Slice Location} & 3 & \tnote{$\ddagger$} & \makecell[c]{G 0020\\E 1041} & DS & Relative position of the image plane expressed in $mm$. \\

            \makecell[r]{Pixel Spacing} & 1C & $\zeta$ & \makecell[c]{G 0028\\E 0030} & DS & Physical distance in the Patient between the center of each pixel, specified by a numeric pair-adjacent row spacing (delimiter) adjacent column spacing in $mm$. \\

            \makecell[r]{Scan Start \\ Location} & P & \tnote{$\divideontimes$} & \makecell[c]{G 0027\\E 1050} & FL & The start position of the entire scan series, same across all slices. \\

            \makecell[r]{Scan End \\ Location} & P & \tnote{$\divideontimes$} & \makecell[c]{G 0027\\E 1051} & FL & The end position of the entire scan series, same across all slices. \\

            \makecell[r]{Recon Center \\ Coordinates} & $\varPhi$ & \tnote{$\divideontimes$} & \makecell[c]{G 0043\\E 1031} & DS & The $x$, $y$ and $z$ center coordinates of the reconstructed area. \\
            \hline
        \end{tabularx}

        \begin{tablenotes}    
            \footnotesize               
            \item[$\dagger$] The require level defined by DICOM, including Required(1), Optional(3), Conditionally Required (1C) and Private(P).
            \item[$\ddagger$] The value representation defined by DICOM, including Decimal String (DS) and Floating Single(FL).
            \item[$\divideontimes$] Tags not used in distortion algorithm but in data loading and sampling.
        \end{tablenotes}
    \end{threeparttable}
\end{table*}

Among the points in \cref{fig:GeometricModeling}, we define the physical table position as:
\begin{equation}
    P_{table} = \left(P_{table}^{(1)},P_{table}^{(2)}\right)=\left(h+\varphi_2,\frac{S_w}{2}\right)
\end{equation}

The vertical distance between the reconstruction field center and table can be calculated as \cref{eq:VerticalDistance}, all elements in this formula are of physical space rather than pixel space:
\begin{equation}
    d = \mathrm{\Phi}_{vertical}-\left(P_{table}^{\left(1\right)}-r\right)=\mathrm{\Phi}_{vertical}-\left(h+\varphi_2-r\right)
    \label{eq:VerticalDistance}
\end{equation}

The unit of $\zeta$ is $mm/pixel$. We define a valid mask with center $\Theta$ and radius $\lambda$ \cref{eq:ValidMask}, which is used to locate the area without undesirable object imaging. All pixels outside this mask will be override by $\varepsilon$.
\begin{equation}
    \begin{aligned}
        & \Theta = (\Theta_x,\Theta_y)=\left( \frac{S_h}{2}-\frac{d}{\zeta}\times\varphi_1,\frac{S_w}{2} \right) \\
        & \lambda = \frac{r}{\zeta}\times\varphi_1 \\
        & For \quad p \quad in \quad \mathcal{R}_{pixels}: \\
        & \quad if \quad \left\lVert {p-\theta}^2 \right\rVert > \lambda^2: \\
        & \quad \quad p = \varepsilon \\
    \end{aligned}
    \label{eq:ValidMask}
\end{equation}

\begin{figure}[tbp]
    \centering
    \includegraphics[width=\linewidth]{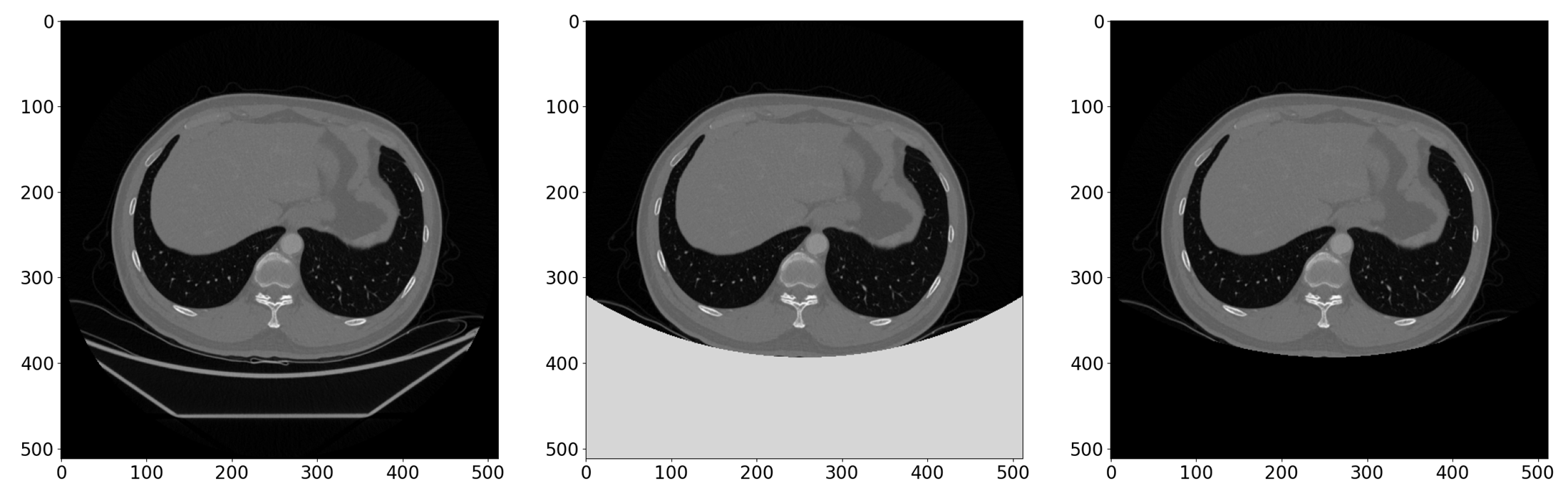}
    \caption{Scan table removal. The subject's body in the left subfigure presents a continuous region resembling an ellipse, and the distorted rectangle below it is the scan table. The metal shell frame of the  table absorbs X-rays strongly, so its shape is clearly shown in the reconstructed image. Our method precisely locates the corresponding area (middle subfigure) and pad it to zero value (right subfigure).}
    \label{fig:ScanTableRemovalEffect}
\end{figure}

\begin{figure}[tbp]
    \centering
    \includegraphics[width=\linewidth]{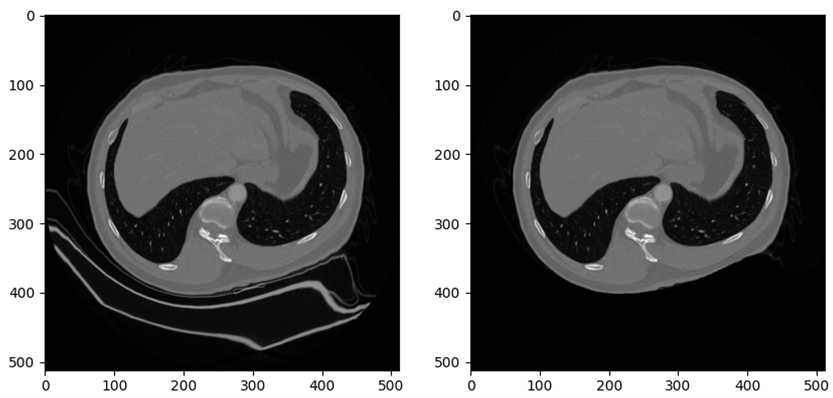}
    \caption{Removal effectiveness on distorted slice. The left subfigure shows that after distortion, the curve of the Scan Table presents a very messy shape, which can introduce significant noise in the feature extraction process. The right one is a slice after the Table Removal operation, where the messy lines have been effectively removed.}
    \label{fig:TableRemovalEffect}
\end{figure}

It is important to note that during the similarity detection process, our objective is to ascertain whether the sample has been excessively distorted. This process is independent of the neural network and maintains an equitable treatment of all pixels, thus the CT Scan Table Removal feature remains inactive during the detection phase. We give several examples of its effect in \cref{fig:ScanTableRemovalEffect} and \cref{fig:TableRemovalEffect}.

The CT Scan Table Removal module necessitates specific DICOM metadata or analogous parameters for its operation, which are generally not available in most public datasets (formatted as NIFTI). Consequently, the application of this module is constrained. Within the scope of our proposed research, this module is not deemed essential. Its utilization is anticipated to enhance the accuracy of subsequent neural networks; however, the accuracy expectations should remain satisfactory even in its absence.

\subsection{Similarity-Guided Hyperparameters Search}

The proposed distortion involves two key parameters (i.e. $A,\omega$) that are specifically designed to control the degree of augmentation. Increasing the value of parameter $A$ results in a higher degree of distortion. Similarly, elevating the value of parameter $\omega$ causes the proposed distortion mapping to resemble a rotational affine more closely. While the intensity of the distortion can be intuitively assessed through visualization, this method does not provide a basis for estimating the accuracy of end-to-end predictions. Hence, we proceed from the assumption that excessive augmentation leads to the destruction of image features, which hinders the neural network’s ability to discern effective features, thereby complicating the training process. We employ SIFT (Scale-Invariant Feature Transform) \citep{Lowe790410, Lowe2004} and ORB (Oriented FAST and Rotated BRIEF) \citep{Rublee6126544} to gauge the similarity of images pre- and post-augmentation \citep{Bansal2021}. At lower augmentation intensities, the sample undergoes minimal changes, allowing the similarity algorithm to identify a greater number of corresponding points between the pre- and post-augmentation samples. However, as the features become compromised, the number of corresponding points diminishes. The performance of similarity detection significantly exceeds that of deep neural networks and remains acceptable when executed on a CPU.

This indicates that we can utilize these metrics to quantify the level of feature degradation. An effective augmentation technique should strike a balance between enhancing the dataset’s information entropy and preserving the learnability of the samples for the neural network. Typically employed in image search, matching, and alignment \citep{Chalom6417053, Csapo6553060}, we pioneer their application in the context of setting augmentation hyperparameters (i.e. $A,\omega$). Our improved training strategy with similarity guide is shown in \cref{algo:Similarity-Guided Hyperparameters Search}. Under the auspices of the Similarity Guide, the FineTune process is streamlined to conduct its search within the $K_{10\%}$ parameter spectrum.
\begin{algorithm}[tbp]
    \fontsize{8pt}{10pt}\selectfont
    \caption{Similarity-Guided Hyperparameters Search}
    \label{algo:Similarity-Guided Hyperparameters Search}
    \begin{minipage}{0.7\linewidth}
        \begin{algorithmic}
            \State Similarity Guideline:
            \State $S\left(I_A,I_W\right)={Sim}_{SIRF}\left(I_A,I_W\right)+{Sim}_{ORB}\left(I_A,I_W\right)$
            \State $K_{10\%}=\left\{\left(I_A,I_W\right) \mid S\left(I_A,I_W\right) \geq t_{\approx90\%}, \left(I_A,I_W\right)\in\left[0.25,12\right]\right\}$
            \State Pretrain:
            \State $W_{pre} \gets Pretrain(W_0,D_{Train})$
            \State Fine-Tune:
            \State $W_k\gets Train\left(W_{pre},D_{Train},\ k\right),\ \forall k\in K_{10\%}$ \Comment{With Distortion and Table Removal}
            \State $L_k\gets Validation(W_k,D_{Val},\ k),\ \forall k\in K_{10\%}$ \Comment{With Table Removal}
            \State ${I_A}^\ast,{I_W}^\ast=K^\ast \gets {\arg\min}_{k \in K_{10\%}}L_k$
        \end{algorithmic}
    \end{minipage}
\end{algorithm}

\subsection{Efficiency}
\label{sec:Efficiency}

The components of the method we have developed are designed to function independently of the neural network. These components can be executed concurrently with neural network computations during the data preprocessing stage, enhancing overall efficiency. Initializing the control point matrix $\mathcal{R}_{Smap}$ has a space complexity of $O(\delta^2)$, and it merely requires memory allocation with almost no additional time cost. Each sample necessitates an individual computation for every control point. The $\mathcal{R}_{Tmap}$ is derived by iterating over all control points. The time complexity is linearly dependent on the number of control points, expressed as $O(\delta^2)$. The computation for each control point entails two conversions between Cartesian and polar coordinate systems, in addition to a polar angle mapping procedure. These operations can be efficiently processed by CPUs.

For triangulation, the Bowyer-Watson based Delaunay method \cite{REBAY1993125} is a prominent technique, exhibiting an average time complexity of $O(nlogn)$. In this context, $n$ represents the total count of pixel points. Given that CT cross-sections commonly feature an initial reconstruction of $512 \times 512$ pixels, this step is recognized as a computationally intensive, pointwise operation. Next, affine mapping point calculations need to be performed for each sub-triangle. We use the affine estimation method provided by Scipy \cite{2020SciPyNMeth}, and its maximum computational complexity arises from the SVD calculation, which is $O(\delta^3)$. Upon the completion of this step, we will have established the point-to-point mapping matrix. For each pixel in the output matrix $\mathcal{R}_{distorted}$ , the corresponding source pixel coordinates are retrieved from the mapping matrix (\cref{eq:PiecewiseAffine}), and sampling is conducted with these coordinates as the center from the source matrix. To mitigate distortion artifacts, bicubic interpolation is utilized during the sampling process. This interpolation technique exhibits a computational complexity of $O(n^2)$.

In summary, the computational overhead of the method we have proposed is primarily associated with the determination of the affine mapping matrix $\mathcal{H}$ and the piecewise affine mapping sampling process. By reducing the resolution $\delta$ of the control point matrix, we can decrease the computational time required for the affine mapping matrix. Similarly, lowering the resolution $\mathcal{S}_h \times \mathcal{S}_w$ of the source matrix can reduce the duration of the sampling execution.

\subsection{IRB Approval}

The dataset is provided by Department of Gastrointestinal Surgery, Shanghai General Hospital. The dataset's details are shown in \cref{table:GastricCancerDatasetOverview} The hospital’s experts labeled the slice containing the largest gastric cancer area for each patient. The research is under the approval from Shanghai General Hospital Institutional Review Board (No. [2024] 032). The Approval Letter is available if required.

\begin{table}[tbp]
    \fontsize{8pt}{10pt}\selectfont
    \raggedright
    \caption{The overview of the dataset collected from clinical practice.}
    \label{table:GastricCancerDatasetOverview}

    \begin{tabularx}{0.5\linewidth}{
        l
        >{\raggedright\arraybackslash}X
        }
        \hline
        \textbf{Item} & \textbf{Describe}\\
        \hline

        Source & Shanghai General Hospital \\
        Date Span & 2014.12.26 $\sim$ 2021.09.18 \\
        Manufacturer & GE MEDICAL SYSTEMS \\
        Model & Revolution CT \\
        Slice Thickness & 0.625mm / 1.25mm / 5mm \\
        Software Versions & sles\_hde3.5 $\sim$ revo\_ct\_21b.32 \\
        Filter & Body \\

        \hline

    \end{tabularx}
\end{table}

\section{Experiments}

\subsection{Data Collection}

In order to validate the efficacy of our augmentation method, we compiled CT scan sequences from 895 patients diagnosed with gastric cancer through clinical practice. A gastrointestinal surgeon identified the slice with the largest gastric cancer lesion area for each patient’s scan, while a radiologist provided detailed cancerous region annotations on this slice, establishing a pixel-level segmentation benchmark. The gastrointestinal surgeon conducted a review of the annotations and made necessary revisions. This approach ensures that only one slice is annotated regardless of the total number of slices in the CT scan sequence. Through a collaborative effort, the radiologist and gastrointestinal surgeon achieved consensus on all scan sequences and segmentations. Following a cleaning process, which excluded cases with \textbf{1)} severe artifacts, \textbf{2)} unsatisfactory imaging, \textbf{3)} unavailable pathological results, \textbf{4)} incomplete lesion coverage, \textbf{5)}indeterminable maximum cross-sectional area due to small lesion size, \textbf{6)} patient or family objections, and \textbf{7)} unavailability of DICOM metadata, we selected 689 cases from the initial 895 for further research.

The dataset incorporated DICOM files with standardized metadata, while the annotation data is preserved separately in nrrd format. The nrrd files contained the original section coordinates, which are utilized to correlate with the DICOM sequences and ascertain the specific section for each annotation. The 895 image sequences collected encompassed reconstruction layer thicknesses of 0.625mm, 1.25mm, and 5mm, with no uniform standard for slice spacing. To ensure consistency, all sequences are resampled to a voxel size of $1mm \times 1mm \times 1mm$, facilitating subsequent training processes.

Additionally, we utilized publicly available large-scale datasets to validate the effectiveness of our proposed method, BraTS\citep{Menze6975210} and CT-ORG\citep{Rister2020, ye2023}.

\subsection{Metrics and Baselines}

We use three pixel-level metrics, Dice, Recall, and Precision, to determine the model’s accuracy in identifying lesions or tissues. The computation formulas for the three metrics are presented below.
\begin{equation}
    \begin{aligned}
        & Dice = \frac{2 \times TP}{2 \times TP + FP + FN} \\
        & Recall = \frac{TP}{TP + FN} \\
        & Precision = \frac{TP}{TP + FP} \\
    \end{aligned}
\end{equation}
where $TP$ represents the number of true positive pixels, $FP$ represents the number of false positive pixels, and $FN$ represents the number of false negative pixels.

Among the techniques employed for data augmentation in interpretable medical image volume sequences, rotation enhancement stands out as one of the most extensively applied and reliable methods \cite{pmlrXEdgeConv}. In terms of neural networks, there are currently two main architechtures, which are the convolutional structure and the Transformer structure. For the public dataset, we have incorporated the SwinUMamba model, which exhibits significant potential, to assess the sensitivity of our method to state space models. At present, the DICOM metadata for most commonly utilized public datasets is not openly available, which constrains the thoroughness of evaluation processes.

\subsection{Similarity Guide Results}
\label{sec:SimilarityGuideResults}

Our primary research goal is to develop a data augmentation technique that is interpretable, does not excessively alter samples, and adheres to clinical norms. The interpretability of the method cannot be assessed through end-to-end training alone. Initially, we employ the proposed image similarity algorithm to quantitatively evaluate whether the samples have been overly distorted by our method. A rapid decline in similarity at certain thresholds would indicate excessive distortion, suggesting lower interpretability as the augmented samples may not be readily comprehensible to clinical professionals.

\begin{figure}[tp]
    \centering
    \subfigure[SIFT]{
        \centering
        \includegraphics[width=0.22\linewidth]{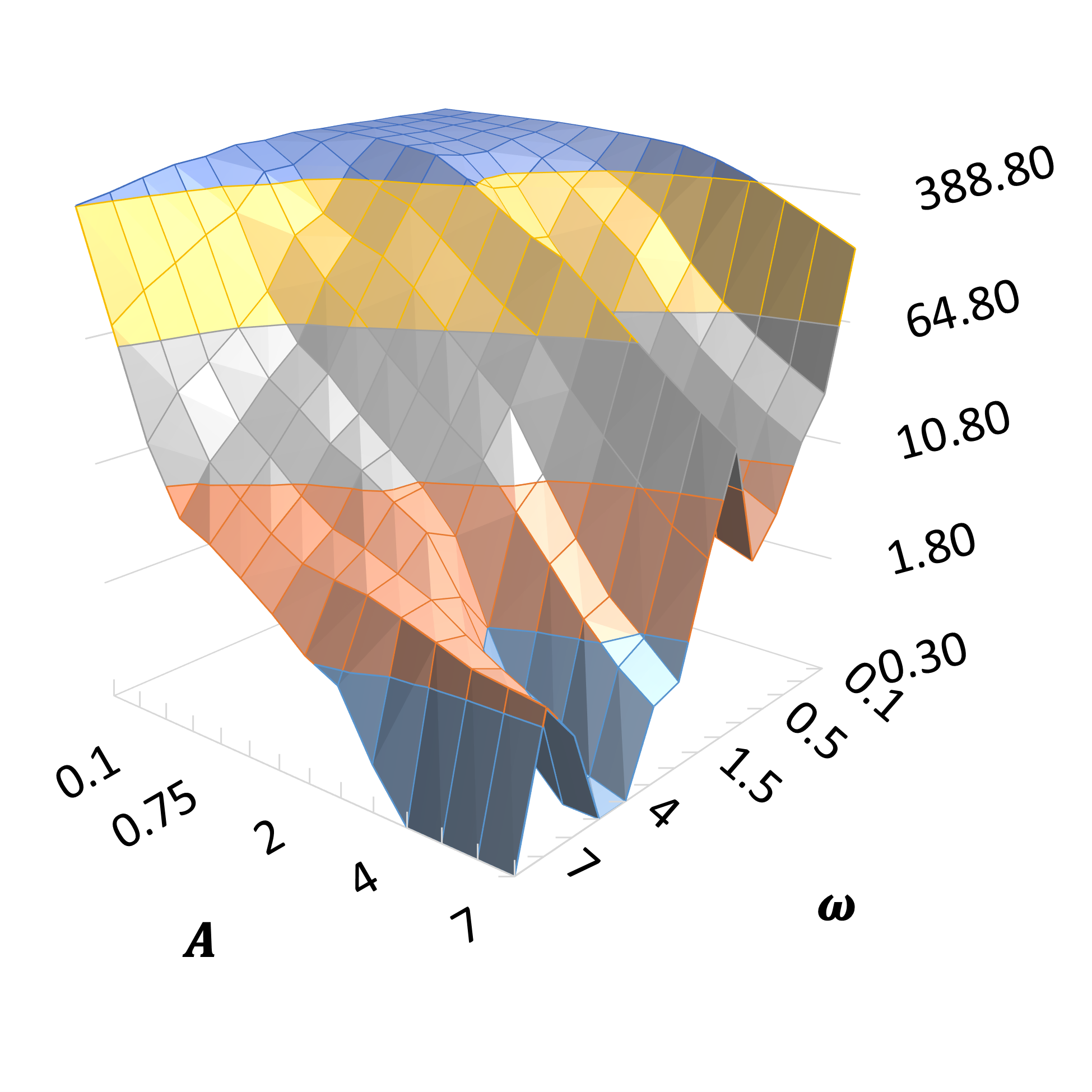}
        \label{fig:SimilarityGuide_Results_SIFT}
    }
    \subfigure[ORB]{
        \centering
        \includegraphics[width=0.22\linewidth]{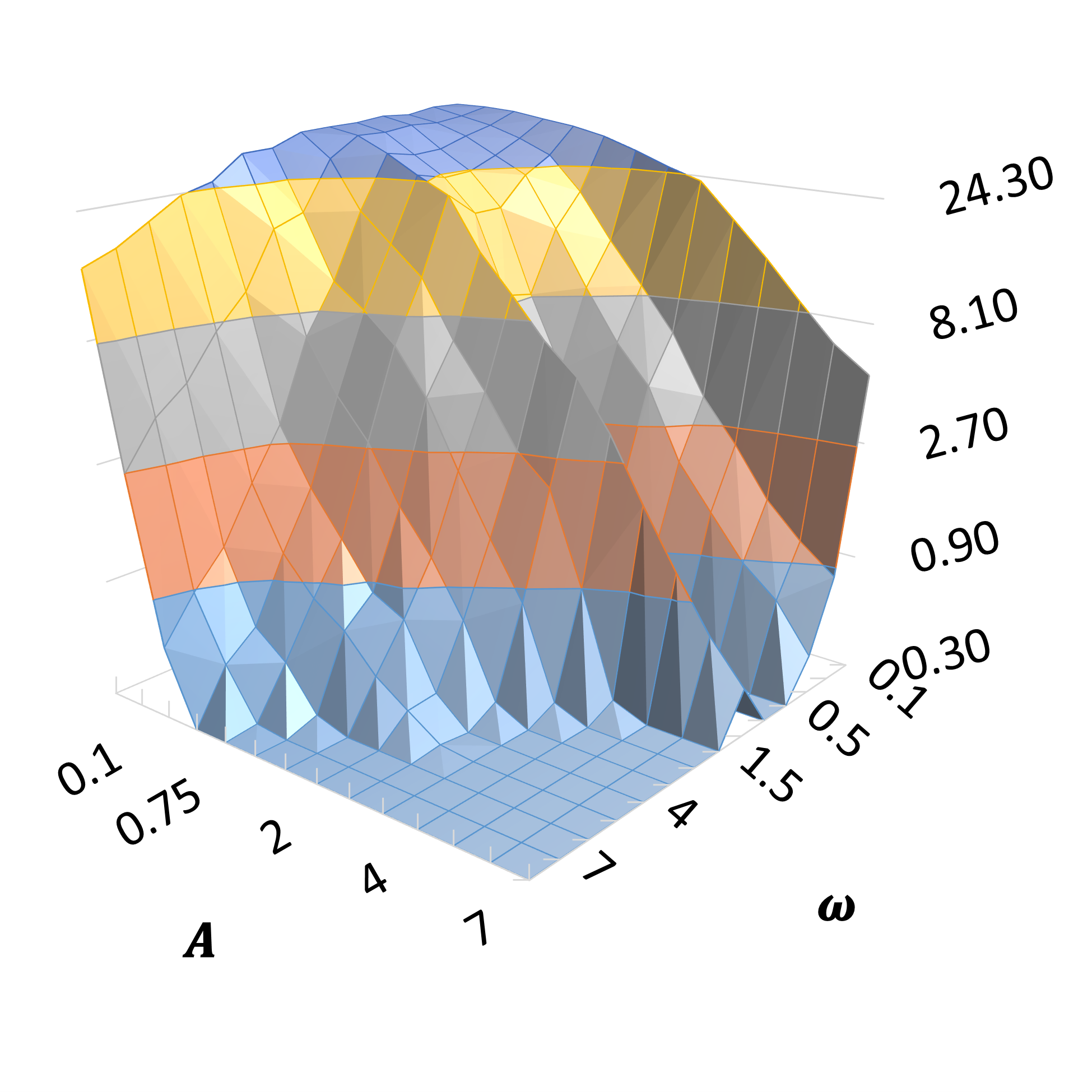}
        \label{fig:SimilarityGuide_Results_ORB}
    }
    \subfigure[Segment Acc.]{
        \centering
        \includegraphics[width=0.22\linewidth]{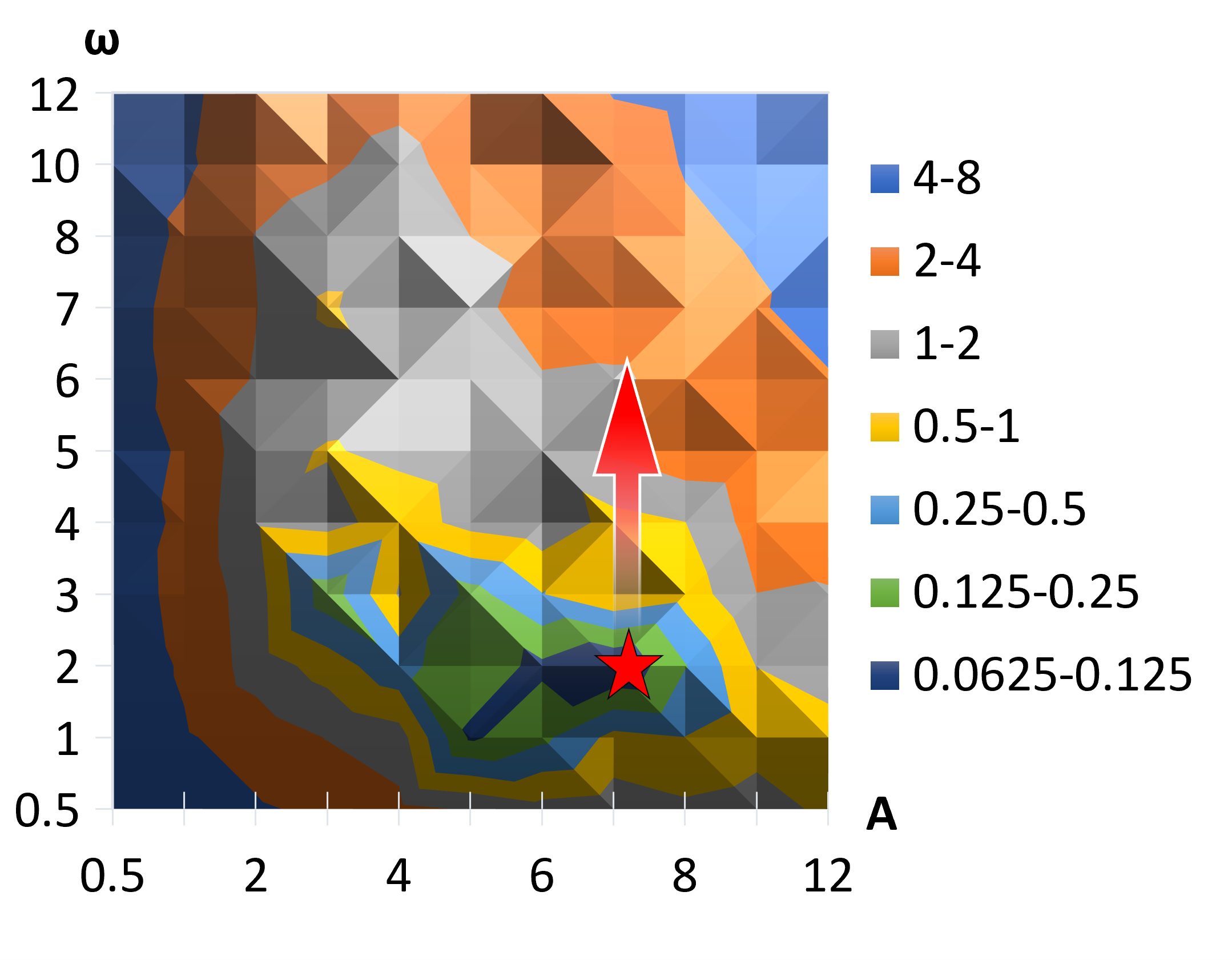}
        \label{fig:SimilarityGuide_Results_SegAcc}
    }
    \subfigure[Correlation]{
        \centering
        \includegraphics[width=0.22\linewidth]{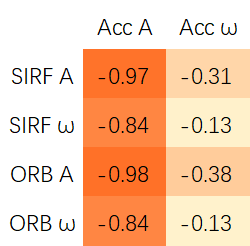}
        \label{fig:SimilarityGuide_Results_Correlation}
    }
    \caption{Similarity guide results. Relations between the Count of Feature Points Matched Successfully and Actual Segmentation Accuracy (\cref{fig:SimilarityGuide_Results_SegAcc}). The feature points are extracted by SIRF (\cref{fig:SimilarityGuide_Results_SIFT}) and ORB (\cref{fig:SimilarityGuide_Results_ORB}) where the Z-axis uses a logarithmic scale to more stably observe the level of correlation. And when $\omega$ exceeds $2$, a significant decline in correlation is observed; in contrast, the effect of $A$ on the correlation is relatively mild. \cref{fig:SimilarityGuide_Results_SegAcc} also shows the same trend (red arrow). \cref{fig:SimilarityGuide_Results_Correlation} is the Pearson correlation coefficient heatmap. A significant negative correlation is observed between similarity and the achievable accuracy.}
    \label{fig:SimilarityGuide_Results}
\end{figure}

The results show that with a variety of parameters employed for augmentation, there is a consistent similarity between the pre- and post-augmentation samples. This initially suggests that our proposed method does not encounter any uncontrollable divergences or singularities. Furthermore, there exists a parameter range that is relatively smooth, within which the method introduces minimal distortion to the images, making it more straightforward for the similarity detection algorithm to perform feature point matching. We observe that the number of successful SIFT pairings starts to plummet with $a\approx3$,\ $f\approx1.5$ \cref{fig:SimilarityGuide_Results}. The Pearson correlation coefficient between model accuracy and two similarity metrics are $\rho_{SIRF-Acc}=-0.625$ and $\rho_{ORB-Acc}=-0.618$.

\begin{figure}[tp]
    \centering
    \subfigure[Similarity Metric]{
        \centering
        \includegraphics[width=0.45\linewidth]{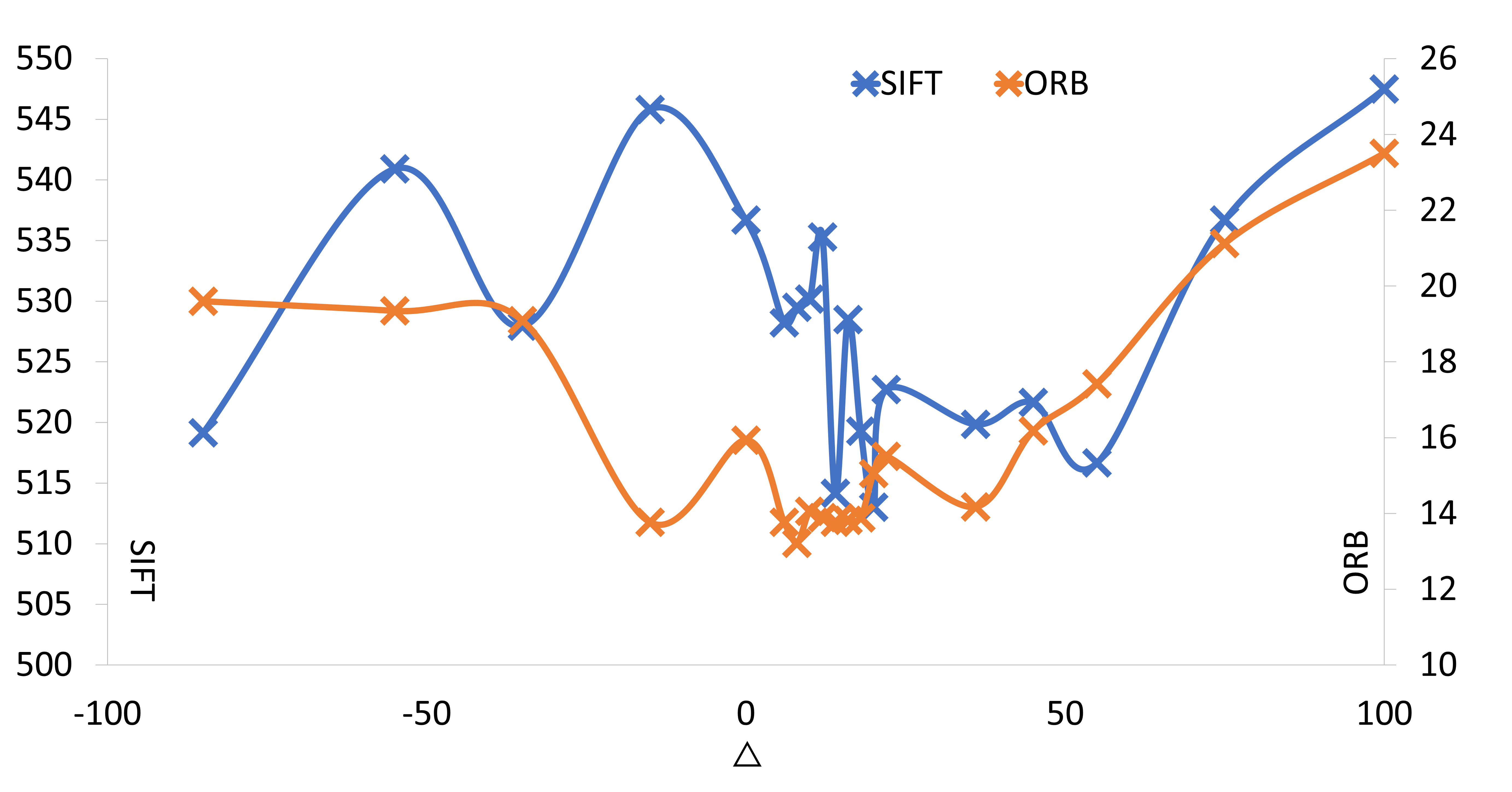}
    }
    \subfigure[Slice Gap on Coronal Section]{
        \centering
        \includegraphics[width=0.45\linewidth]{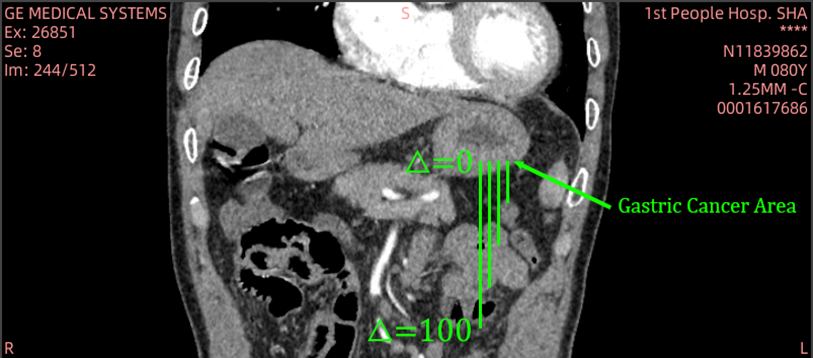}
    }
    \caption{Similarity distribution around labelled slices. The vertical axis in the first and second subfigure correspond to SIFT and ORB value. The number of similarity detection points shows an upward trend when $\triangle\rightarrow u, u\notin stomach\approx\left[-20,20\right]$. The similarity results indicate a scarcity of distinctive features in the axial slice where the stomach is located. This aligns with anatomical realities: the heart and lungs are characterized by a higher density of blood vessels and bronchi, while the mid-to-lower abdomen contains a diverse array of organs.}
    \label{fig:AdjacentSimilarityDistribution}
\end{figure}

Given that our annotations are focused on gastric lesions, we have conducted an in-depth examination of the similarity levels across upper abdomen. Here, we aimed to assist in a more comprehensive determination of the optimal distortion intensity \cref{fig:AdjacentSimilarityDistribution}. The similarity criterion is described in \cref{eq:Similarity_Criterion}. A higher $N_\triangle$ indicates a greater complexity of textures in the respective axial position, enabling the extraction of a larger number of feature points. 
\begin{equation}
    \begin{aligned}
        & N_\triangle = \left\{\sum_{i\in I} S i m\left(\mathcal{R}_\triangle,\ Distortion\left(\mathcal{R}_\triangle, A=j, W=i\right)\right) \mid j \in I\right\} \\
        & I=\left\{0.5,1,2,3,4,5,6,7,8\right\} \\
    \end{aligned}
    \label{eq:Similarity_Criterion}
\end{equation}
where $\triangle$ represent the distance on $Y$ axis between labelled slices and target slices.

While similarity detection is initially employed to ascertain a general range for the augmentation parameters, \cref{sec:Ablation_Param} involved training each parameter from scratch. This approach is taken to validate the reliability of the ideal parameter range that is more efficiently derived through similarity detection.

\subsection{Segmentation Evaluations}

We examine our method's effectiveness the latest segmentation models, MedNext\citep{Roy101007} and SwinUMamba\citep{liu2024swinumamba}, combined with the large dataset provided by SA-Med2D\citep{ye2023}. The results are presented in \cref{tab:Segmentation_CTORG,tab:Segmentation_BraTS}.
\begin{table}[tbp]
    \fontsize{8pt}{10pt}\selectfont
    \centering
    \caption{Accuracy on the large CT-ORG\citep{Rister2020} dataset. Our method is able to help the latest neural networks to achieve higher accuracy with limited training samples, specially useful in medical scenarios. This dataset is a typical representative of CT modality.}
    \label{tab:Segmentation_CTORG}
    \begin{tabularx}{\linewidth}{
        c
        c
        c
        >{\centering\arraybackslash}X
        >{\centering\arraybackslash}X
        >{\centering\arraybackslash}X
        >{\centering\arraybackslash}X
        >{\centering\arraybackslash}X
        >{\centering\arraybackslash}X
        >{\centering\arraybackslash}X
        }
        \hline

        \multirow{2}{*}{Model} & \multirow{2}{*}{Implementation} & \multirow{2}{*}{Metric} & \multicolumn{6}{c}{Classes} & \multirow{2}{*}{Average} \\
        \cline{4-9}
        & & & backgound & bladder & bone & kidney & lung & liver \\
        \hline

        \multirow{7}{*}{MedNext} & \multirow{3}{*}{Rotation $\pm90^\circ$} & Dice & 96.39 & 65.53 & 3.21 & 19.32 & 66.26 & 86.13 & 56.14 \\
        & & Precision & 96.01 & 62.91 & 12.70 & 76.04 & 67.88 & 78.92 & 65.74 \\
        & & Recall & 96.77 & 68.37 &  1.84 & 11.07 & 64.72 & 94.80 & 56.26 \\

        \cline{2-10}

        & \multirow{3}{*}{With Ours} & Dice & 96.85 & 78.54 & 48.55 & 36.08 & 71.19 & 90.04 & 70.21 \\
        & & Precision & 95.60 & 79.93 & 88.49 & 78.35 & 82.75 & 89.58 & 85.78 \\
        & & Recall & 98.13 & 77.19 & 33.45 & 23.44 & 62.47 & 90.50 & 64.20 \\

        \cline{2-10}

        & \multicolumn{2}{r}{\textbf{Average Improve}} & \textbf{+0.47} & \textbf{+12.95} & \textbf{+50.91} & \textbf{+10.48} & \textbf{+5.85} & \textbf{+3.42} & \textbf{+14.01} \\

        \hline

        \multirow{7}{*}{SwinUMamba} & \multirow{3}{*}{Rotation $\pm90^\circ$} & Dice & 96.85 & 71.44 & 1.39 & 21.38 & 71.84 & 92.3 & 59.21 \\
        & & Precision & 95.27 & 80.27 & 47.50 & 70.38 & 78.02 & 91.40 & 77.14 \\
        & & Recall & 98.49 & 64.36 & 0.71 & 12.61 & 66.56 & 93.34 & 56.01 \\

        \cline{2-10}

        & \multirow{3}{*}{With Ours} & Dice & 97.27 & 74.52 & 58.87 & 32.81 & 73.57 & 90.23 & 71.21 \\
        & & Precision & 96.04 & 73.63 & 82.58 & 70.40 & 86.26 & 91.52 & 83.41 \\
        & & Recall & 98.53 & 75.44 & 45.74 & 21.39 & 64.13 & 88.98 & 65.70 \\

        \cline{2-10}

        & \multicolumn{2}{r}{\textbf{Average Improve}} & \textbf{+0.41} & \textbf{+2.51} & \textbf{+45.86} & \textbf{+6.74} & \textbf{+2.51} & -2.12 & \textbf{+9.32} \\

        \hline
    \end{tabularx}
\end{table}
\begin{table}[tbp]
    \fontsize{8pt}{10pt}\selectfont
    \centering
    \caption{The similar experiment with \cref{tab:Segmentation_CTORG}, but is on the BraTS\citep{Menze6975210} dataset, and trained with SwinUMamba. This dataset is a typical representative of MR modality.}
    \label{tab:Segmentation_BraTS}
    \begin{tabularx}{\linewidth}{
        c
        c
        c
        >{\centering\arraybackslash}X
        >{\centering\arraybackslash}X
        >{\centering\arraybackslash}X
        >{\centering\arraybackslash}X
        c
        }
        \hline

        \multirow{2}{*}{Challenge Year} & \multirow{2}{*}{Implementation} & \multirow{2}{*}{Metric} & \multicolumn{4}{c}{Classes} & \multirow{2}{*}{Average} \\
        \cline{4-7}
        & & & backgound & non\_enhancing\_tumor & enhancing\_tumor & edema \\
        \hline

        \multirow{7}{*}{BraTS 2020} & \multirow{3}{*}{Rotation $\pm90^\circ$} & Dice & 99.06 & 37.69 & 13.75 & 26.93 & 44.36  \\ 
        ~ & ~ & Precision & 98.43 & 42.70 & 23.00 & 33.51 & 49.41  \\ 
        ~ & ~ & Recall & 99.69 & 33.74 & 9.80 & 22.51 & 41.44  \\ 

        \cline{2-8}

        ~ & \multirow{3}{*}{WIth Ours} & Dice & 99.14 & 36.40 & 16.84 & 35.78 & 47.04  \\ 
        ~ & ~ & Precision & 98.69 & 36.15 & 27.32 & 42.36 & 51.13  \\ 
        ~ & ~ & Recall & 99.59 & 36.64 & 12.17 & 30.97 & 44.84  \\ 

        \cline{2-8}

        & \multicolumn{2}{r}{\textbf{Average Improve}} & \textbf{+0.08}  & -1.65 & \textbf{+3.26}  & \textbf{+8.72}  & \textbf{+2.60} \\

        \hline

        \multirow{7}{*}{BraTS 2021} & \multirow{3}{*}{Rotation $\pm90^\circ$} & Dice  & 99.03 & 35.24 & 1.53 & 30.76 & 41.64  \\ 
        ~ & ~ & Precision  & 98.52 & 39.7 & 9.94 & 38.29 & 46.61  \\ 
        ~ & ~ & Recall  & 99.54 & 31.68 & 0.83 & 25.71 & 39.44  \\ 

        \cline{2-8}

        ~ & \multirow{3}{*}{WIth Ours} & Dice  & 99.04 & 33.09 & 3.17 & 36.06 & 42.84  \\ 
        ~ & ~ & Precision  & 98.49 & 48.76 & 19.92 & 39.28 & 51.61  \\ 
        ~ & ~ & Recall  & 99.61 & 25.04 & 1.72 & 33.32 & 39.92  \\ 

        \cline{2-8}

        & \multicolumn{2}{r}{\textbf{Average Improve}} & \textbf{+0.02}  & \textbf{+0.09}  & \textbf{+4.17}  & \textbf{+4.63}  & \textbf{+2.23}   \\ 

        \hline
    
    \end{tabularx}
\end{table}

In \cref{tab:Segmentation_Imp_Overview}, we evaluated our method on several traditional and widely-used neural networks, training on our private gastric cancer dataset. The dataset allows us to examine the effectiveness of the proposed Metadata-Driven Scan Table Removal and Distortion Augmentation. The results show an improvement on segmentation accuracy across multiple neural-network-based segmentation frameworks, which are selected from the most representative ones in recent computer vision researches. Most frameworks can steady gain higher value on major metrics without the extra samples. Our method has demonstrated a consistent enhancement in Dice scores across various experiments, excluding the Poolformer model. The improvement is evident even when distortion is utilized without the Table Removal feature. However, the incremental gain in accuracy when Table Removal is activated is relatively minor, suggesting its impact may not be as significant.
\begin{table}[tbp]
    \fontsize{8pt}{10pt}\selectfont
    \centering
    \caption{The improvements on our private gastric cancer datasets with only one annotations per scan, which is a challenging scenario for traditional frameworks. These models are famous and widely-used, so may further demonstrated good universality.}
    \label{tab:Segmentation_Imp_Overview}
    \begin{tabularx}{\linewidth}{
        >{\centering\arraybackslash}X
        >{\centering\arraybackslash}X
        >{\centering\arraybackslash}X
        >{\centering\arraybackslash}X
        >{\centering\arraybackslash}X
        >{\centering\arraybackslash}X
        >{\centering\arraybackslash}X
        >{\centering\arraybackslash}X
        >{\centering\arraybackslash}X
        }
        \hline

        Family & Model & Criterion & w/o aug\tnote{1} & rotate $\pm 45^\circ$ & rotate $\pm 90^\circ$ & rotate $\pm 180^\circ$ & w/o Table Removal & Ours \\
        
        \hline

        \multirow{12}{*}{Conv-Based} & \multirow{4}{*}{Resnet50} & mIoU & 72.52 & 78.32 & 79.49 & 78.58 & \underline{81.27} & \textbf{82.57} \\

        & & mDice & 81.28 & 86.28 & 87.20 & 86.48 & \underline{88.56} & \textbf{89.52} \\

        & & mRecall & 82.39 & 87.34 & 88.67 & \textbf{91.53} & 88.90 & \underline{91.37} \\

        & & mPrecision & 83.81 & 87.04 & 87.80 & 85.87 & \textbf{89.07} & \underline{89.05} \\

        \cline{2-9}

        & \multirow{4}{*}{ConvNext} & mIoU & 63.47 & 71.99 & 75.2 & 75.00 & \underline{77.46} & \textbf{79.64} \\

        & & mDice & 71.72 & 80.77 & 83.75 & 83.51 & \underline{85.58} & \textbf{87.32} \\

        & & mRecall & 77.36 & 82.26 & 84.36 & 86.95 & \underline{88.68} & \textbf{90.37} \\

        & & mPrecision & 75.54 & 82.57 & 83.75 & 82.83 & \underline{85.94} & \textbf{87.16} \\

        \cline{2-9}

        & \multirow{4}{*}{SegNeXt} & mIoU & 64.02 & 72.46 & 73.04 & 74.16 & \underline{76.37} & \textbf{76.61} \\

        & & mDice & 72.37 & 81.23 & 81.77 & 82.76 & \underline{84.67} & \textbf{84.88} \\

        & & mRecall & 70.31 & 83.81 & 86.20 & 85.98 & \underline{87.06} & \textbf{89.39} \\

        & & mPrecision & 77.46 & 80.73 & 82.54 & 83.60 & \underline{84.34} & \textbf{85.57} \\

        \hline

        \multirow{16}{*}{Trans-Based} & \multirow{4}{*}{MAE} & mIoU & 67.81  & 71.73  & 70.00  & 68.90 & \textbf{75.92} & \underline{75.04} \\

        & & mDice & 76.63  & 80.56  & 78.89  & 77.81 & \textbf{84.30} & \underline{83.55} \\

        & & mRecall & 85.79  & 85.80  & 82.25  & 84.59 & \textbf{87.84} & \underline{87.09} \\

        & & mPrecision & 78.27  & 77.44  & 78.66  & 73.90 & \underline{81.86} & \textbf{82.70} \\

        \cline{2-9}

        & \multirow{4}{*}{Poolformer} & mIoU & 68.08  & 76.16  & 76.64  & \underline{76.84} & \textbf{78.60} & 76.63 \\

        & & mDice & 76.88  & 84.50  & 84.91  & \underline{85.07} & \textbf{86.50} & 84.90 \\

        & & mRecall & 76.26  & 88.00  & \underline{89.89}  & \textbf{90.48} & 88.98 & 89.00 \\

        & & mPrecision & 79.06  & 83.41  & \textbf{84.37}  & 83.63 & \underline{84.33} & 83.62 \\

        \cline{2-9}

        & \multirow{4}{*}{Segformer} & mIoU & 67.68  & 77.62  & 77.58  & 74.04 &\underline{81.19} & \textbf{82.94} \\

        & & mDice & 76.48  & 85.72  & 85.68  & 82.67 & \underline{88.51} & \textbf{89.78} \\

        & & mRecall & 80.61  & 92.99  & 91.87  & 86.55 & \underline{93.40} & \textbf{95.15} \\

        & & mPrecision & 79.50  & 84.27  & 83.68  & 79.84 & \underline{85.14} & \textbf{90.87} \\

        \cline{2-9}

        & \multirow{4}{*}{Swin Trans. V2} & mIoU & 65.72 & 72.97 & 73.87 & 73.73 & \underline{76.34} & \textbf{79.81} \\

        & & mDice & 74.31  & 81.70  & 82.51  & 82.39 & \underline{84.65} & \textbf{87.46} \\

        & & mRecall & 74.41  & 83.58  & 85.14  & 85.00 & \underline{87.96} & \textbf{91.43} \\

        & & mPrecision & 77.76  & 80.03  & 80.34  & 80.16 & \underline{83.48} & \textbf{86.71} \\

        \hline
    \end{tabularx}

    \begin{tablenotes}
        \footnotesize
        \item[1] Trainings with the minimum preprocesses required for model training (i.e. loading, type convert, resize).
    \end{tablenotes}

\end{table}

In the context of MAE, the exclusion of the Table Removal process yields superior accuracy. This is predominantly due to the masked reconstruction phase inherent in MAE’s learning mechanism, which affords equal consideration to all pixels within the slice, thereby not treating the CT Scan Table as noise. The removal of the Table leads to a reduction in image information entropy, which simplifies the learning task for MAE and diminishes the learning space. As a result, the model’s ultimate accuracy is compromised.

\begin{figure}[tp]
    \centering
    \includegraphics[width=\linewidth]{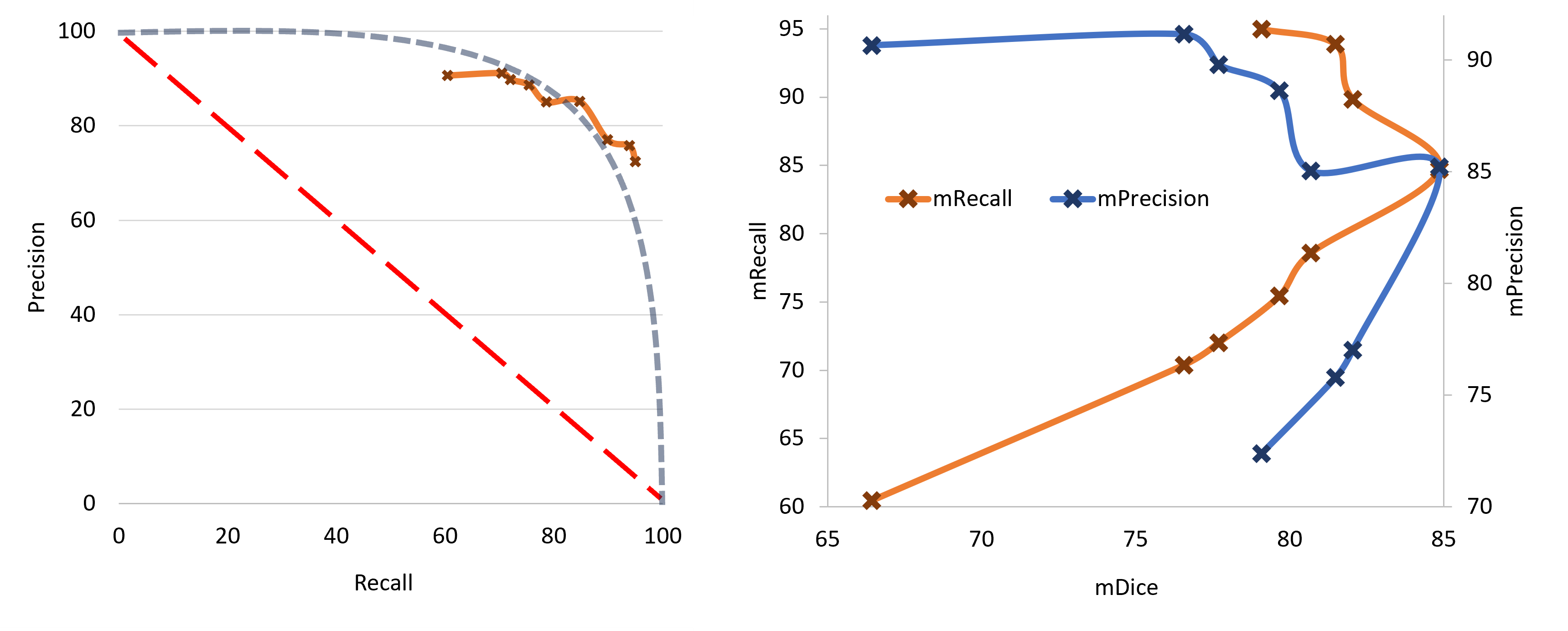}
    \caption{Segmentation metrics. In the left subfigure, the orange dots represent actual data points, and the gray dashed line indicates the estimated PR curve. A curve closer to upper right corner represents better. The right subfigure illustrates the counterbalance between accuracy criterions (mDice) and positive pixel criterions (mRecall and mPrecision). The results demonstrate that our model can produce effective classification.}
    \label{fig:Segmentation_Metrics}
\end{figure}

\cref{fig:Segmentation_Metrics} illustrate the Precision-recall Curve. Due to limited computational resources, we focused on calculating the most dynamically changing segmentgain higher value on  of the PR curve.

\subsection{Ablation of Distortion Parameter}
\label{sec:Ablation_Param}

\begin{figure}[tp]
    \centering
    \includegraphics[width=\linewidth]{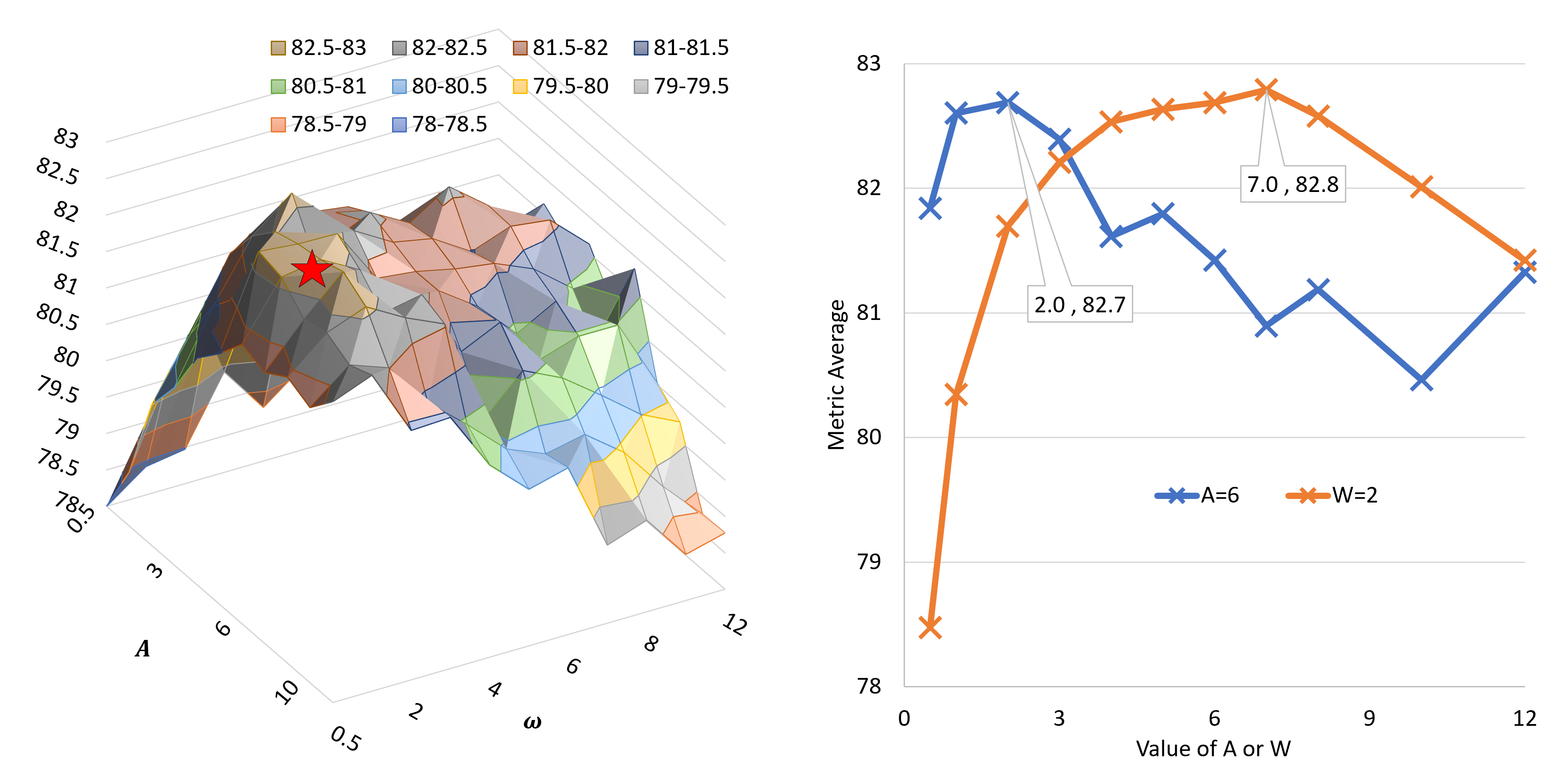}
    \caption{Parameter search on $A$ and $\omega$. Too small or too large $A$ and $\omega$ will lead to reduced accuracy, and their distribution is relatively moderate, reflecting the high stability of the algorithm under different parameters. The most effective setting seems to be around $A=6,\omega=2$.These results are very close to the results of the similarity calculation. This proves that the similarity calculation can effectively estimate the reasonable augmentation intensity range with much lower computational complexity.}
    \label{fig:ParameterSearch}
\end{figure}

We conducted ablation experiments on parameters $A$ and $\omega$ respectively as is shown in \cref{fig:ParameterSearch}. Overall, our model can achieve acceptable results within a large range of augmentation parameters, and the effects of adjacent parameters tend to be similar, which is manifested as a smoother surface in the accuracy distribution map. This feature ensures its ease of use, after all, researchers always tend to prefer a plug-and-play module rather than excessive parameter tuning. 

When $A$ or $\omega$ becomes too large, it will instead reduce the segmentation accuracy of the model. This is because excessive distortion will cause adjacent slice regions to be overly distorted, and the neural network cannot extract effective pixel features from these regions. Furthermore, larger $A,\omega$ parameters usually require larger $\delta$ to maintain the resolution of affine interpolation. The overall complexity of distortion can be approximated as $O(\delta^2)$. If hardware constraints only allow the algorithm to operate with a smaller $\delta$, the augmented image may contain increased fragmentation.

When comparing the results to \cref{sec:SimilarityGuideResults}, there appears to be a correlation between the number of matching points identified by the similarity algorithm and the accuracy of the end-to-end prediction. The neural network's accuracy starts to falter rather than improve when the augmentation intensity with $A>7, \omega>3$, and the number of successful pairings during similarity calculation starts to plummet with $a\approx3$,\ $f\approx1.5$. Noted that during training, we use \cref{eq:RandomAW} to randomly determine $A, \omega$ for each training batch, while use constant value during similarity calculation. Consequently, the mean of the sampling distribution for random parameters during the training phase can serve as a benchmark for comparison with the parameter values applied in the similarity detection process. Based on these observations, it is evident that there is a similarity between the distributions in question. The similarity detection and the neural network’s predictions for slices exhibit a rapid decline in segmentation accuracy under similar augmentation intensities. This implies that the proposed similarity detection algorithm has the potential to accurately estimate the optimal parameter range without the prerequisite of actual neural network training, thereby substantially reducing the computational expense associated with parameter adjustment.

\subsection{Computation Complexity}

\begin{figure}[tbp]
    \centering
    \includegraphics[width=\linewidth]{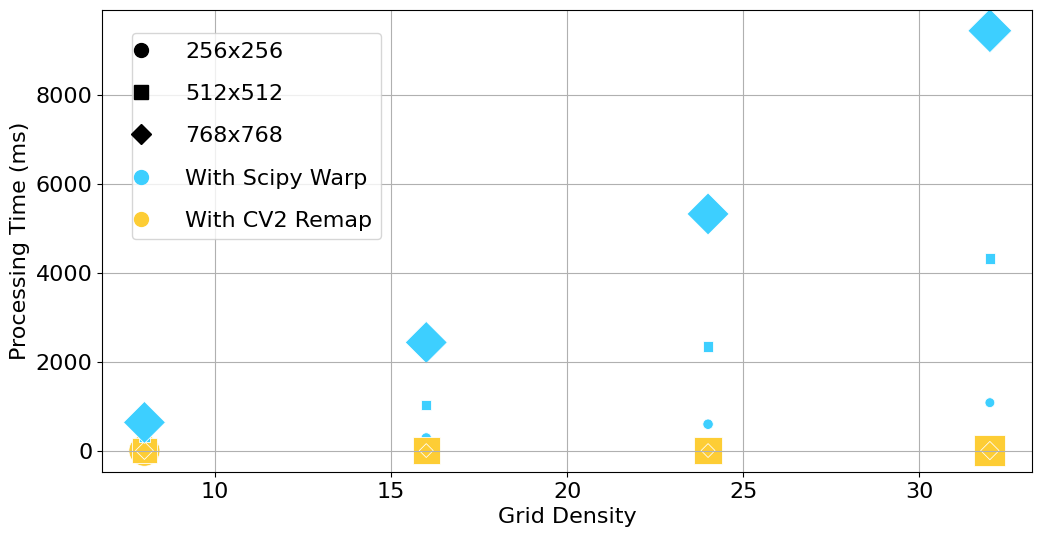}
    \caption{Computational Overhead Analysis. Distinct shapes are utilized to denote varying sizes $n$ of the source pixel matrices. Different colors are assigned to indicate the selection of post-sampling implementation options. Additionally, the size of the shapes corresponds to the relative memory footprint. In summary, the efficiency of Scipy is inferior, whereas cv2 consistently demonstrates millisecond-level responsiveness across all configurations.}
    \label{fig:AugCompuationOverhead}
\end{figure}

In line with the theoretical analysis provided in \cref{sec:Efficiency}, our implementation identifies the determination of the affine mapping matrix and the mapping sampling process as the primary computational bottlenecks, which are correlated with the quantity of control points and source pixels, respectively. We conducted ablation experiments on these two factors, and the findings are presented in \cref{fig:AugCompuationOverhead}.

The experimental findings indicate a positive correlation between the number of control points (Grid Density) and processing time. Additionally, an increase in the resolution of the source matrix results in higher memory consumption. When employing OpenCV for mapping sampling instead of Scipy\cite{2020SciPyNMeth}, the study observed substantially reduced processing times and lower memory usage. As the computation method for the mapping matrix remained unchanged, utilizing the same piecewise affine mapping technique with Delaunay triangulation, it is inferred that the primary time overhead is attributed to the resampling of the source matrix.

The aforementioned data is obtained through the process of independent data augmentation execution. In order to better stick to the practical application scenarios, we conducted a further analysis to ascertain whether data preprocessing acts as a bottleneck in the computational speed of neural networks during model training, thereby potentially causing periods of GPU idleness. The results are presented in \cref{tab:GPU_Idle}. The GPU exhibits minimal idleness during the training process. This is attributed to the CPU in completing the preprocessing of subsequent batch of data in advance of the completion of the current batch’s neural network computations, which further indicating the efficiency of the proposed method.
\begin{table}
    \raggedright
    \caption{Computational Overhead Analysis. The table lists the average time taken for each step in the data augmentation process. The time unit is in milliseconds. The results indicate that the data augmentation process is not a bottleneck in the training process. }
    \label{tab:GPU_Idle}
    \begin{tabular}{rcc}
        \toprule
        Model & Batch Time & GPU Idle\\
        \midrule
        ConvNeXt & 200 & 2.2 \\
        MAE & 271 & 2.0 \\
        Poolformer & 142 & 3.5 \\
        ResNet50 & 259 & 1.7 \\
        Segformer & 239 & 2.6 \\
        SegNeXt & 123 & 3.2 \\
        SwinT. V2 & 194 & 3.9 \\
        \bottomrule
    \end{tabular}
\end{table}

\section{Discussions}

To our knowledge, our approach is one of the few instances within the Medical Imaging domain that employs image similarity metrics for neural network hyperparameter search \citep{app12083754}. Similar to the meta-data-driven Scan Table Removal technique, it offers significantly higher throughput compared to approaches relying solely on deep learning, while maintaining comparable accuracy. To implement these methods, researchers need to delve into a deeper understanding of medical imaging sequences. Our research suggests that while the approach to determining control points for segmented affine mapping is intuitive, this approach may inadvertently constrain the diversity of outcomes. Merely designating the area adjacent to the spine as the focal point for distortion may not yield optimal results. Given the current capability in the academic community to automate the localization of numerous human organs, tissues, and structures, applying a regional-level segmented affine mapping individually to each target instance based on these localization results could potentially enhance the robustness of the outcomes and increase the variety of the samples generated.

In the context of X-Ray imaging, it is common to obtain projections of slices exclusively on the coronal plane. This limitation implies that the method introduced in this research is not readily applicable to Chest X-Rays (CxR). Given the greater prevalence of X-Rays compared to CT scans, a technique that circumvents the need for slice modeling would enhance the scope of augmentation.

Our objective is to capitalise on the distinct attributes of medical 3D Volume imagery in contrast to traditional visual imaging or point cloud 3D imagery. We intend to develop a highly persuasive augmentation approach that harnesses the inherent benefits of medical 3D Volume images. In the realm of healthcare, it is imperative that not only neural networks but every component of the implementation garners sufficient credibility within clinical contexts. This level of trust is a fundamental requirement for the effective fusion of artificial intelligence with medical practices.

\section{Conclusion}
In this paper, we propose an augment method for scan series using polar-sine-based piecewise affine distortion. This method is able to generate any number of virtual samples from an existing scan sequence while ensuring that the relative anatomical structures of the human body are not severely altered, thereby enhancing the learning capability of downstream neural networks. The method is easy to deploy in today's mainstream deep learning frameworks and is compatible with most medical radiologic imaging data containing Slice-Wise dimension. Experiments have proven that this method can provide significant accuracy improvements on various types of deep-learning-based segmentation models.

\section*{Acknowledgement}

The work is supported in part by National Natural Science Foundation of China (Grant Nos. 61572325) and the Shanghai Key Technology Project (19DZ1208903).

I am profoundly grateful to my advisor (Prof. Chen) for generously providing the essential research environment, computational resources, and data resources that are crucial for the advancement of this study. I also extend my heartfelt thanks to my girlfriend (M. Eng. Chen) for her steadfast support and companionship, which greatly contributed to my perseverance during the most tough time.

\section*{Data Availability}
The implementation code used in this research is available online: https://github.com/MGAMZ/PSBPD.

\printcredits

\bibliographystyle{cas-model2-names}
\bibliography{PSBPD.bib}

\end{document}